\documentclass[10pt,twocolumn,letterpaper]{article}

\usepackage{cvpr}              % To produce the CAMERA-READY version
\usepackage{graphicx}
\usepackage{amsmath}
\usepackage{amssymb}
\usepackage{booktabs}
\usepackage{tabularx}
\usepackage{adjustbox}
\usepackage{float}
\usepackage{wrapfig}
\usepackage{multirow}
\usepackage[accsupp]{axessibility}

\usepackage[pagebackref,breaklinks,colorlinks]{hyperref}

\usepackage[capitalize]{cleveref}
\crefname{section}{Sec.}{Secs.}
\Crefname{section}{Section}{Sections}
\Crefname{table}{Table}{s}
\crefname{}{Tab.}{Tabs.}

\def\cvprPaperID{6492} % *** Enter the CVPR Paper ID here
\def\confName{CVPR}
\def\confYear{2022}

\begin{document}

\title{Parametric Scattering Networks}

% \author{\centering%
%     Shanel Gauthier$^{1,3*}$ \hspace{-10}
%     \qquad Benjamin Th\'{e}rien$^{2,3,4*}$ \hspace{-10}
%     \qquad Laurent Als\`{e}ne-Racicot$^{1,3}$ \hspace{-10}
%     \qquad Muawiz Chaudhary $^{2,3}$\\%
%     }

\author{\parbox{\textwidth}{\centering
    Shanel Gauthier$^{1,2}\;$\thanks{Equal contributions. $^{\dagger}$Equal senior author contribution. This research was partially funded by NSERC CGS-M~[\emph{S.G.,L.A.}] and URA~[\emph{B.T.}] scholarhips; NSERC Discovery Grant RGPIN-2021-04104.~[\emph{E.B.}]; IVADO PRF Grant [\emph{I.R.,E.B.,G.W.}]; and CIFAR AI Chairs [\emph{I.R.,G.W.}]. We acknowledge resources provided by Compute Canada and Calcul Quebec. The content is solely the responsibility of the authors and does not necessarily represent the official views of funding agencies. Correspondence to: \texttt{eugene.belilovsky@concordia.ca, meickenberg@flatironinstitute.org, guy.wolf@umontreal.ca}} \hspace{-10pt}
    \qquad Benjamin Th\'{e}rien$^{2,3,4*}$ \hspace{-10pt}
    \qquad Laurent Als\`{e}ne-Racicot$^{1,2}$ \hspace{-10pt}
    \qquad Muawiz Chaudhary $^{3,2}$\\
    Irina Rish$^{1,2}$
    \qquad Eugene Belilovsky$^{3,2 \dagger}$
    \qquad Michael Eickenberg $^{5 \dagger}$
    \qquad Guy Wolf$^{1,2 \dagger}$} \vspace{5pt}\\
$^1$ Universit\'{e} de Montr\'{e}al; $^2$ Mila -- Quebec AI Institute; $^3$ Concordia University, Montreal, QC, Canada \\
$^4$Waterloo University, Waterloo, ON, Canada;$^5$ Flatiron Institute, New York, NY, USA\\ 
% {\texttt{\small\{shanel.gauthier, laurent.alsene-racicot,irina.rish,eugene.belilovsky,guy.wolf\}@umontreal.ca}}\\{\texttt{\small btherien@uwaterloo.ca muawiz.chaudhary@mila.quebec meickenberg@flatironinstitute.org}}
}

% \and

\maketitle
%%%%%%%%% ABSTRACT
\begin{abstract}\vspace{-4pt}
   The wavelet scattering transform creates geometric invariants and deformation stability. In multiple signal domains, it has been shown to yield more discriminative representations compared to other non-learned representations and to outperform learned representations in certain tasks, particularly on limited labeled data and highly structured signals. The wavelet filters used in the scattering transform are typically selected to create a tight frame via a parameterized mother wavelet. In this work, we investigate whether this standard wavelet filterbank construction is optimal. Focusing on Morlet wavelets, we propose to learn the scales, orientations, and aspect ratios of the filters to produce problem-specific parameterizations of the scattering transform. We show that our learned versions of the scattering transform yield significant performance gains in small-sample classification settings over the standard scattering transform. Moreover, our empirical results suggest that traditional filterbank constructions may not always be necessary for scattering transforms to extract effective representations.
\end{abstract}
\vspace{-12pt}
%%%%%%%%% BODY TEXT
\section{Introduction}
\label{sec:introduction}

The scattering transform, proposed in~\cite{mallat2012group}, is a cascade of wavelets and complex modulus nonlinearities, which can be seen as a convolutional neural network (CNN) with fixed, predetermined filters. This construction can be used to build representations with geometric invariants and is shown to be stable to deformations. It has been demonstrated to yield impressive results on problems involving highly structured signals \cite{bruna2013invariant,oyallon2013generic, anden2014deep, sifre2014rigid, hirn2015quantum, hirn2017wavelet, eickenberg2018solid, anden2019joint, sinz2020wavelet, perlmutter2020geometric}, outperforming a number of other classic signal processing techniques. Since scattering transforms are instantiations of CNNs, they have been studied as mathematical models for understanding the impressive success of CNNs in image classification \cite{bruna2013invariant,mallat2016understanding}. As discussed in~\cite{bruna2013invariant}, first-order scattering coefficients are similar to SIFT descriptors \cite{SIFT}, and higher-order scattering can provide insight into the information added with depth \cite{mallat2016understanding}. Moreover, theoretical and empirical study of information encoded in scattering networks indicates that they often promote linear separability, which leads to effective representations for downstream classification tasks \cite{bruna2013invariant,oyallon2017scaling,anden2015joint,eickenberg2018solid}.

Scattering-based models have been shown to be useful in several applications involving scarcely annotated or limited labeled data~\cite{bruna2013invariant,sifre2013rotation,oyallon2018replearning,eickenberg2018solid}. Indeed, most breakthroughs in deep learning in general, and CNNs in particular, involve significant effort in collecting massive amounts of well-annotated data to be used when training deep overparameterized networks. While big data is becoming increasingly prevalent, there are numerous applications where the task of annotating more than a small number of samples is infeasible, giving rise to increasing interest in small-sample learning tasks and deep-learning approaches towards them~\cite{ barz2020deep,learnfewsamples, Worrall_2017_CVPR}. Recent work has shown that, in image classification, state-of-the-art results can be achieved by hybrid networks that harness the scattering transform as their early layers followed by learned layers based on a wide residual network architecture~\cite{oyallon2018replearning}. Here, we further advance this research avenue by proposing to use the scattering paradigm not only as fixed preprocessing layers in a concatenated architecture, but also as a parametric prior to learn filters in a CNN. This allows us to also shed light on whether the  standard wavelet construction \cite{mallat1999wavelet} is an optimal approach for building filterbanks from a mother-wavelet for discriminative tasks.% This way, we permit early-layer filters to adapt to data in a way that guarantees a meaningful filter footprint.

%Specifically we formulate a parametric scattering, whose parameters we can adapt to the problem.  
Recall that the scattering construction is based on complex wavelets, generated from a mother wavelet via dilations and rotations, aimed to cover the frequency plane while having the capacity to encode informative variability in input signals~\cite{bruna2013invariant}. Further, discrete parameterization and indexing of these operations (i.e., by dilation scaling or rotation angle) have traditionally been carefully constructed to ensure the resulting filter bank forms an efficient tight frame \cite{mallat1999wavelet,mallat2012group} with well-established energy preservation properties. On the other hand, it has been observed that the first layers of convolutional networks resemble wavelets but may not necessarily form a tight frame \cite{krizhevsky2012imagenet}. The question then arises: is it necessary to use the standard wavelet filterbank construction? Here, we relax the standard construction by considering another alternative where a small number of wavelet parameters used to create the wavelet filterbanks are optimized for the task at hand.
%, while maintaining the filterbank construction and use of mother wavelet. 

%We also relax both constructions to allow data-driven learning (i.e., via backpropagation) of the wavelet parameters used in scattering layers of hybrid architectures \footnote{The Code is available on  \url{https://github.com/psn-iclr-submission/iclr_anon}}.

To our knowledge, this is the first work that aims to learn the wavelet filters of scattering networks in 2D signals.
%, and more generally to pose strict parametric priors on filters learned in early layers of 2D convolutional networks. 
Related work and the empirical protocol are summarized in Sec.~\ref{sec:parametric_scattering_networks}. and  Sec.~\ref{sec:exp-setup} respectively.  In Sec.~\ref{section:converging}, we compare scattering parameterizations obtained from optimizing over different datasets. %explore the different filter construction schemes by comparing the wavelet filter parameterizations they produce when optimized over different datasets. 
In Sec.~\ref{section:deformation}, we evaluate the robustness of our parametric scattering networks to deformation. In Sec.~\ref{sec:smalldata}, we demonstrate the advantages of our approach in limited labeled data settings and study the adaptation of the wavelet parameters toward a supervised task. In Sec.~\ref{sec:unsup}, we investigate the adaptation of the parametrized scattering using an unsupervised objective. Finally, in Sec.~\ref{sec:computation} we evaluate the computational and memory complexity of our hybrid networks. Further technical details appear in appendices, and code accompanying the work is available at \url{https://github.com/bentherien/parametricScatteringNetworks}.

\begin{figure*}
    \centering
    \includegraphics[width=0.85\linewidth]{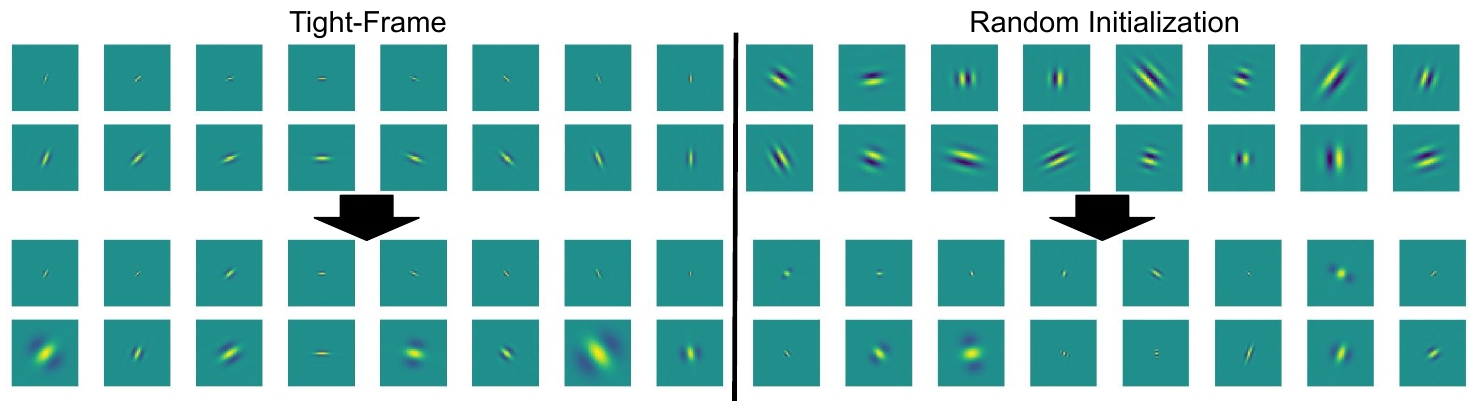}
    \caption{\textbf{Initialized wavelet filters pre and post-training.} Real part of Morlet wavelet filters initialized with \textit{tight-frame} (left) and  \textit{random} (right) schemes before (top) and after (bottom) training. The filters were optimized on the entire CIFAR-10 training set with linear model. We use the Morlet canonical wavelet parameterization. For the tight-frame filters, we observe substantial changes in both scale and aspect ratio. On the other hand, all random filters undergo major changes in orientation and scale. }
    \label{fig:filters-cifar-real-before-kymatio}
    \vspace{-10pt}
\end{figure*}
\vspace{-5pt}

\section{Related work}
\label{sec:relatedwork}
Learning useful representations from little training data \cite{learnfewsamples} is arduous and a reality in a variety of domains such as in biomedicine and healthcare. 
Recent works have tried to tackle this problem. Lezama et al.~\cite{lezama2018ole} replace the categorical cross-entropy loss with a geometric loss called Orthogonal Low-rank Embedding (OLÉ) to reduce the intra-class variance and enforce inter-class margins. Barz and Denzler~\cite{barz2020deep} also propose to replace the categorical cross-entropy loss, but this time with the cosine loss function in order to decrease overfitting in the small-sample classification settings. The cosine loss function, as opposed to the softmax function used with cross-entropy, does not push the logits of the true class to infinity.
%as explained in \cite{szegedy2016rethinking}.
Other methods show promise by incorporating prior knowledge into the model~\cite{Gens_2014_NIPS, Jacobsen_2016_CVPR, bruintjes2021vipriors, Kayhan_2020_CVPR, brigato2021tune}. Oyallon et al.~\cite{oyallon2018replearning} introduce hybrid networks where the scattering transform with fixed wavelets was shown to be an effective replacement for early layers of learned convolutional networks on a wide residual network architecture. 
%Similarly, \cite{raghu2019transfusion} propose to initialize the filters of the first layer of a CNN as synthetic Gabor filters where they are not constrained to remain Gabor-like during the training. 
Cotter and Kingsbury~\cite{cotter2019learnable} also propose a hybrid network called a learnable ScatterNet, where learning layers are intermixed between the scattering orders, unlike our work where only a few parameters governing the wavelet construction are modified.  Ulicny et al.~\cite{8902831} propose Harmonic Networks (HN), a hybrid network consisting of fixed Discrete Cosine Transform filters combined with learnable weights in CNNs.

Related to our work, adding learnable components to existing wavelet-based representations has been considered in a number of recent works in the context of time-series~\cite{balestriero2018spline, seydoux2020clustering, cosentino2020learnable, balestriero2020interpretable}. Balestriero et al.~\cite{balestriero2018spline} and Seydoux et al.~\cite{seydoux2020clustering} learn a spline-parametrized mother wavelet for 1D problems. Similarly, Cosentino and Aazhang~\cite{cosentino2020learnable} parametrized the group transform in the context of time-series data. Our work, alternatively, focuses on 2D problems and maintains the canonical Morlet wavelet parameterization, but allows deviation from a tight-frame filter bank.
%The concept of bringing learnabibility into scattering networks have also been explored recently on times-series applications \cite{balestriero2018spline, seydoux2020clustering, cosentino2020learnable, balestriero2020interpre}. 

%In the context of image classification, similarly to our work, \cite{alekseev2019gabornet} propose the GaborNet where the filters in the first layer of the network are constrained to fit the Gabor function. Scattering transforms are not used as a prior in the GaborNet and the different approaches of Gabor parameters initialization are not studied in their work. 

\section{Parameterization of Scattering Networks}
%\vspace{-0.05in}
\label{sec:parametric_scattering_networks}
We first revisit the formulation of traditional scattering convolution networks in Sec.~\ref{sec:scatteringnetworks} and introduce our parametric scattering transform in Sec.~\ref{sec:parametricoptim} and ~\ref{sec:eq}. Finally, Sec.~\ref{sec:init} discusses scattering parameter initialization.

%\vspace{-0.05in}
\subsection{Scattering Networks}
%\vspace{-0.05in}
\label{sec:scatteringnetworks}

For simplicity, we focus here on 2D scattering networks up to their 2nd order. Subsequent orders can be computed by following the same iterative scheme, but have been shown to yield negligible energy \cite{bruna2013invariant}. Given a signal $x(u)$, where $u$ is the spatial position index, we compute the scattering coefficients $S^0x, S^1x, S^2x$, of order 0, 1, and 2 respectively. For an integer $J$, corresponding to the spatial scale of the scattering transform, and assuming an $N \times N$ signal input with one channel, the resulting feature maps are of size $\frac{N}{2^J}\times\frac{N}{2^J}$, with channel sizes varying with the scattering coefficient order (i.e., 1 channel at order 0, $JL$ channels at order 1 and $L^2J(J-1)/2 $ channels at order 2).   

To calculate 0th-order coefficients, we consider a low-pass filter $\phi_J$ with a spatial window of scale $2^J$, such as a Gaussian smoothing function. We then convolve this filter with the signal and downsample by a factor of $2^J$ to obtain $S^0x(u) = x*\phi_J(2^Ju).$
%Three observations can be made at this point. First, this feature map has only one channel. Second, these calculations can be effectively done in the Fourier domain. Third, by using an averaging filter, high frequency information is lost. %information about high frequency is lost. 
Due to the low-pass filtering, high-frequency information is discarded here and is recovered in higher-order coefficients via wavelets introduced as in a filter bank.

\begin{figure*}[t]
    \centering
    \includegraphics[width=0.49\textwidth]{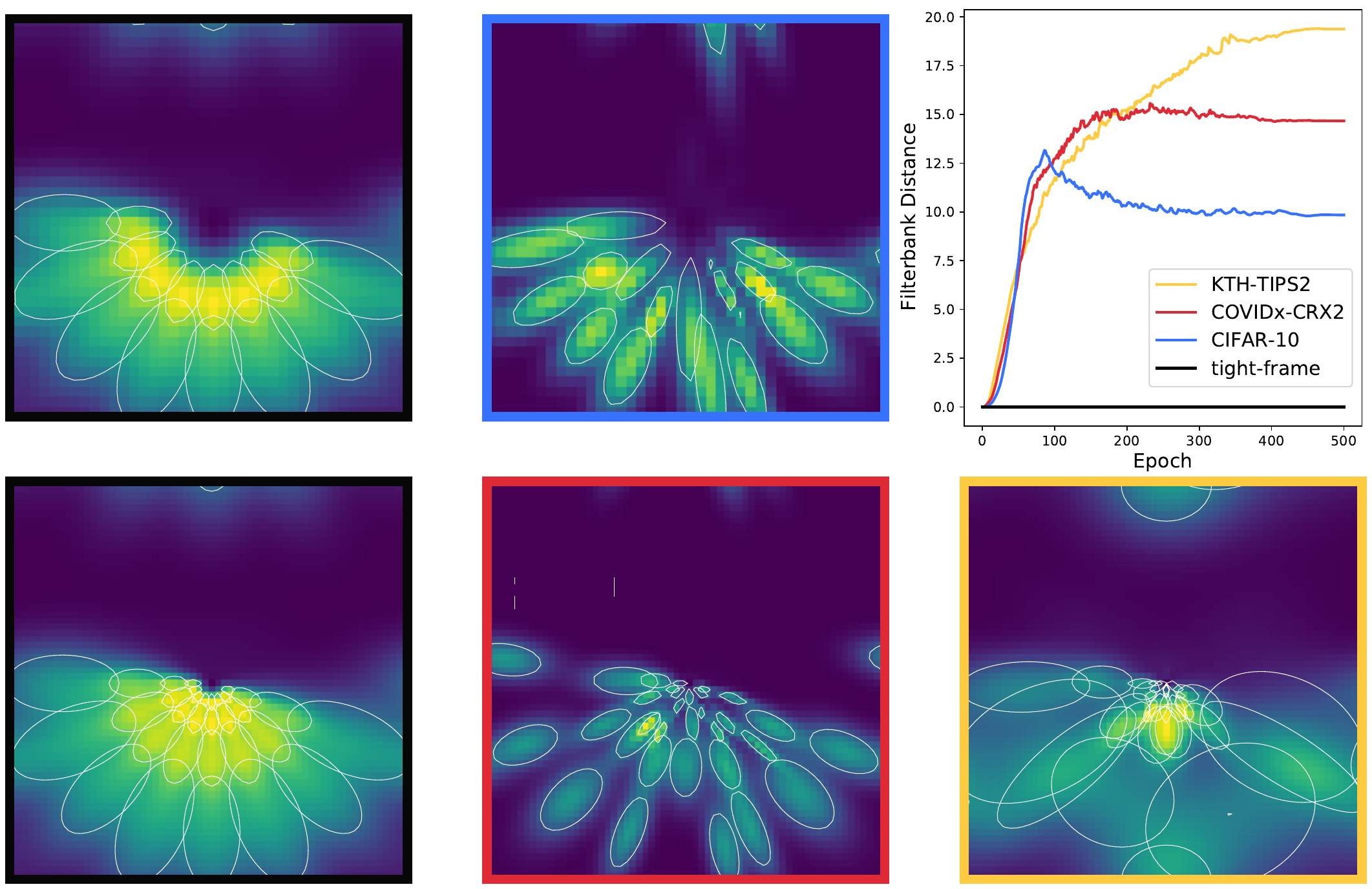} 
    \vspace{-7pt}
    \caption{\textbf{Parametric scattering network learns dataset specific filters.} The graph (top right) shows the \textit{filterbank distance} over epochs as the filters are trained on different datasets. We visualize dataset specific parameterizations of scattering filterbanks (border colors from the legend) in Fourier space.  The x and y axis are the frequency axis. %White contours are drawn around each morlet wavelet for clarity. 
   Scattering filters optimized for natural (CIFAR-10) and medical image (COVIDx CRX2) become more orientation-selective, i.e., thinner in the Fourier domain. On the other hand, filters optimized for texture discrimination (KTH-TIPS2) become less orientation-selective and deviate most from a tight-frame setup.\vspace{-10pt}}
    \label{fig:graphlwp}
\end{figure*}

Morlet wavelets are a typical example of filters used in conjunction with the scattering transform, and are defined as
\vspace{-15pt}
\begin{equation}\label{eq:1}
    \psi_{\sigma, \theta, \xi, \gamma}(u) = e^{-\|D_{\gamma}R_{\theta}(u)\|^2/(2\sigma^2)}(e^{i\xi u'}-\beta),
\end{equation}
where $\beta$ is a normalization constant to ensure wavelets integrate to 0 over the spatial domain, $u' = u_1 \cos \theta + u_2 \sin \theta$, $R_{\theta}$ is the rotation matrix of angle $\theta$ and 
$
D_{\gamma} = \begin{pmatrix}
1 & 0 \\
0 & \gamma
\end{pmatrix}
.
$
The four parameters can be adjusted and are presented in Table~\ref{tab:params}. 
From one wavelet $\psi_{\sigma^\prime, \theta^\prime, \xi^\prime, \gamma^\prime}(u)$, the traditional wavelet filterbank is obtained by dilating it by factors $2^j$, $0\leq j < J$, and rotating by $L$ angles $\theta$ equally spaced over the circle, to get $\{2^{\textrm{-}2j} \psi_{\sigma^\prime, \theta^\prime, \xi^\prime, \gamma^\prime}(2^{\textrm{-}j}R_{\theta}(u))\}$, which is then completed with the lowpass $\phi_J$.  This can be written in terms of the parameters in Table~\ref{tab:params} as $\psi_{2^{j} \sigma^\prime, \theta^\prime - \theta, 2^{\textrm{-}j}\xi^\prime, \gamma^\prime}(u) = \psi(2^{\textrm{-}j}R_{\theta}(u))$. By slight abuse of notations, we use $\psi_\lambda$ here, $\lambda = (\sigma_j, \theta, \xi_j, \gamma_j)$, to denote such wavelets indexed by $\theta$ and $j$. The resulting set of filters is visualized in the frequency domain in Figure \ref{fig:graphlwp}. 

\begin{table}[tb]
\centering
\small
\caption{Canonical Parameters of Morlet wavelet\vspace{-5pt}}
    \begin{tabular}{|c|c||c|c|}
     \hline %\hhline{|-|-||-|-}
     Param  &  Role &  Param  &  Role\\\hline
     % \hhline{|=|=||=|=|}%\hhline{==::==} 
      $\sigma$   & Gaussian window scale & 
      $\theta$   & Global orientation \\
      \hline%\hhline{--||--}
      $\xi$   & Frequency scale & 
      $\gamma$   & Aspect Ratio \\
      \hline %hhline{--||--}
    \end{tabular}
    \label{tab:params}
\vspace{-15pt}
\end{table}

First-order scattering coefficients are calculated by first convolving the input signal with one of the generated complex wavelets (i.e., indexed by the parameters in Table~\ref{tab:params}) and downsampling the resulting filtered signal by the scale factor $2^{j_1}$ of the wavelet chosen. Then, a pointwise complex modulus is used to add nonlinearity, and the resulting real signal is smoothed via a low-pass filter. Finally, another downsampling step is applied, this time by a factor of $2^{J-j_1}$, to obtain an optimally compressed output size. Mathematically, we have 
% $$
% S^1x(\lambda_1,u) = |x*\psi_{\lambda_1}|*\phi_J(2^Ju). 
% $$
\begin{align}
S^1x(\lambda_1,u) = |x*\psi_{\lambda_1}|*\phi_J(2^Ju). 
\end{align}
The resulting feature map has $J \cdot L$ channels, based on the number of wavelets in the generated family. Second-order coefficients are generated similarly, with the addition of another cascade of wavelet transform and modulus operator before the low-pass smoothing, i.e.,
% $$
% S^2x(\lambda_1,\lambda_2,u) = ||x*\psi_{\lambda_1}|*\psi_{\lambda_2}|*\phi_J(2^Ju). 
% $$
\begin{align}
S^2x(\lambda_1,\lambda_2,u) = ||x*\psi_{\lambda_1}|*\psi_{\lambda_2}|*\phi_J(2^Ju). 
\end{align}
Due to the interaction between the bandwidths and frequency supports of first and second order, only coefficients with $j_1<j_2$ have significant energy. Hence, the second-order output yields a feature map with 
$\frac{1}{2}J(J-1)L^2$ channels. 

\subsection{Morlet Canonical Parameterization}
\label{sec:parametricoptim}
%\vspace{-0.05in}
While the wavelet filters are traditionally fixed, we let the network learn the optimal parameters of each wavelet. In other words, we constrain our filters to always be Morlet wavelets by only optimizing the parameters in Table~\ref{tab:params}. We call this approach the Morlet canonical parameterization of the wavelet. To provide such data-driven optimization of scattering parameters, we show, in Appendix \ref{sec:derivative}, that it is possible to backpropagate through this construction. We adapted the Kymatio software package \cite{andreux2020kymatio} to create the learnable scattering network.

%\subsection{ Wavelet Filters Pixel-Wise Optimization}
%\label{sec:pixel}
%Parametric scattering networks constrain the filters to always be Morlet wavelets. As a naive alternative, we relax the constraints and instead optimize the pixels of the wavelets. This alternative is similar to initializing the kernels of a convolutional neural network with Morlet wavelets and let the model optimize the pixels of the kernels without any constraint using backpropagation. 

\subsection{Initialization}
\label{sec:init}
% \vspace{-0.05in}
To evaluate the importance of the standard wavelet construction, we consider two initializations and study their impact on resulting performance in both learned and nonlearned settings. First, the standard wavelet construction follows common implementations of the scattering transform by setting $\sigma_{j,\ell} = 0.8\times 2^j$, $\xi_{j,\ell} = \frac{3\pi}{4}2^{\textrm{-}j}$, and $\gamma_{j,\ell} = \frac{4}{L}$ for $j = 1,\ldots,J$, $\ell = 1,\ldots,L$, while for each $j$, we set $\theta_{j,\ell}$ to be equally spaced on $[0,\pi)$. The construction ensures the resulting filter bank forms an efficient tight frame. Thus, we call this construction the tight-frame initialization. Second, as an alternative, we consider a random initialization where these parameters are sampled as $\sigma_{j,\ell} \sim \log(U[\exp 1,\exp 5])$, $\xi_{j,\ell} \sim U[0.5,1]$, $\gamma_{j,\ell} \sim U[0.5,1.5]$, and $\theta_{j,\ell} \sim U[0,2\pi]$. %\textcolor{red}{Add here some motivation of the latter one.}
That is, orientations are selected uniformly at random on the circle, the filter width $\sigma$ is selected using an exponential distribution across available scales and the spatial frequency $\xi$ is chosen to be in the interval $[0.5, 1]$, which lies in the center of the feasible range between aliasing ($>\pi$) and the fundamental frequency of the signal size ($2\pi / N$ where $N$ is the number of pixels). Finally, we select the \textit{aspect ratio} variable to vary around the spherical setting of $1.0$, with a bias towards stronger orientation selectivity ($0.5$) compared to lesser orientation selectivity ($1.5$).
\vspace{-3pt}

\subsection{Morlet Equivariant Parameterization }
\label{sec:eq}
In the Morlet canonical parameterization approach, the canonical parameters of each filter are learned. As an alternative method, we consider the Morlet equivariant parameterization in which the number of learnable parameters is reduced by a factor L compared to the Morlet canonical parameterization. Each filter per scale is constructed using the same four parameters in Table \ref{tab:params}: $\sigma$, $\xi$, $\gamma$ and $\Theta$. 
%The second filter is built using the same parameters except for the global orientation which is set to $\Theta + \frac{\pi}{L}$.
However, the global orientation of the $L$ filters for each scale are set to be [$\Theta$, $\Theta + \frac{\pi}{L}$, $\Theta + \frac{2\pi}{L}$, \ldots ,  $\Theta + \frac{(L-1)\pi}{L}$]. 
%By construction, the tight-frame filters are equivariant. 

\section{Experiments}
\label{sec:exp-setup}
\label{sec:results}
Our empirical evaluations are based on three image datasets: CIFAR-10,  COVIDx CRX-2, and KTH-TIPS2. CIFAR-10 and KTH-TIPS2 are natural image and texture recognition datasets, correspondingly. They are often used as general-purpose benchmarks in similar image analysis settings \cite{glico,sifre2013rotation}. COVIDx CRX-2 is a dataset of X-ray scans for COVID-19 diagnosis; its use here demonstrates the viability of our parametric scattering approach in practice, e.g., in medical imaging applications.

We evaluate the use of the parametrized scattering with two common models. In the first case, we consider the scattering as feeding into a simple linear model (denoted LL). The LL configurations are used to evaluate the linear separability of the obtained scattering representations and have the added benefit of providing a more interpretable model. In the second case, we take the approach of \cite{oyallon2018replearning} and  consider the scattering as the first stage of a deeper CNN, specifically a Wide Residual Network (WRN)~\cite{wideresnet}. The architecture of the WRN hybrid is described in more detail in Appendix~\ref{appendix-wrn}.

For both models (LL and WRN), we compare learned parametric scattering networks (LS) to fixed ones (S). For learned scattering (LS), we consider two scattering parameterization approaches: \textit{Morlet canonical}, described in Sec.~\ref{sec:parametricoptim} and \textit{Morlet equivariant},  described in Sec.~\ref{sec:eq}. To show the importance of the parametric approach, we also ablate the naive parameterization where all pixels of the wavelets are adapted, which we refer to as a pixel-wise parameterization.
%consider an additional, purposefully naive, approach where we optimize the pixels of the wavelets directly in Fourier space. Parametric scattering networks constrain the filters to always be Morlet wavelets. In the pixel-wise parameterization approach, we relax the constraints and instead optimize the pixels of the Fourier transform of the wavelets. This alternative is similar to initializing the kernels of a CNN with Morlet wavelets and letting the model optimize the pixels of the kernels without any constraint using backpropagation. 
For each scattering architecture, we consider both random and tight-frame (TF) initialization.  The fixed scattering models determined by the TF construction are equivalent to traditional scattering transforms. Finally, we also compare our approach to a fully learned WRN (with no scattering priors) and ResNet-50~\cite{he2016deep} applied directly to input data. We note that the latter is unmodified form its ImageNet architecture %(i.e., not specialized for $32^2$ or $128^2$ images) 
and that we do not initialize it with pre-trained weights. 

Across all scattering configurations, a batch-normalization layer with learnable affine parameters is added after all scattering layers. Classification is performed via a softmax layer yielding the final output. All models are trained using cross-entropy loss, minimized by stochastic gradient descent with momentum of 0.9. Weight decay is applied to the linear model and to the WRN. The learning rate is scheduled according to the one cycle policy~\cite{smith2019super}. %The scheduler's div factor is always set to 25. 
Implementation details specific to each dataset are described in Appendix \ref{appendix-details}. 
We replicate some of the experiments with learnable scattering networks followed by WRN on CIFAR-10, COVIDx-CRX2, and KTH-TIPS2 using the cosine loss function \cite{barz2020deep}. The results are reported in Appendix~\ref{appendix-cosine-loss}.
%, which improves convergence during optimization, especially in the small data regime.%, due to its so-called super convergence (see \cite{smith2019super} for further details).%, while learning rate is tuned.}

\subsection{Exploring Dataset-specific Parameterizations}
\label{section:converging}

\begin{figure*}
    \centering
    \includegraphics[width=0.27\linewidth]{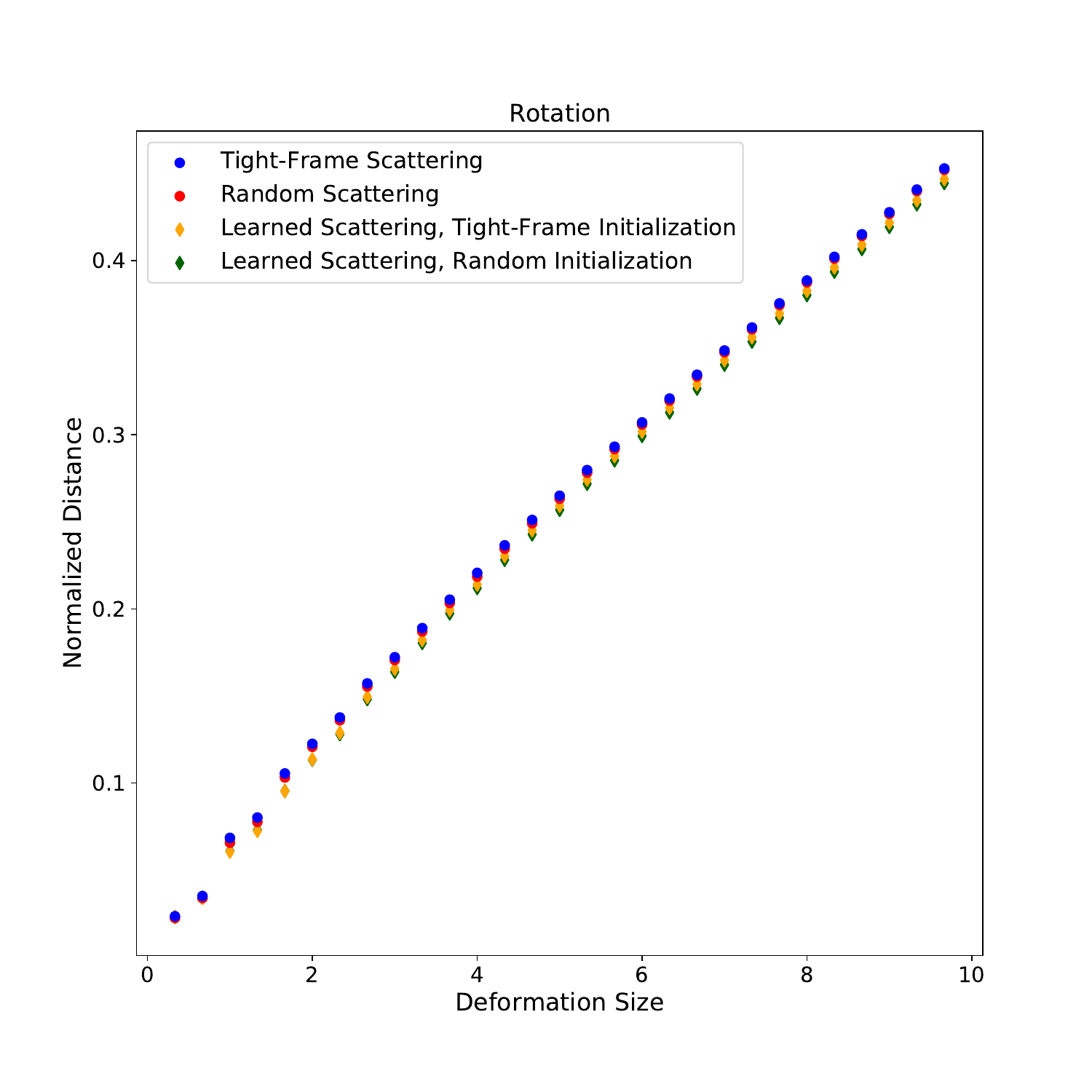}
    \includegraphics[width=0.27\linewidth]{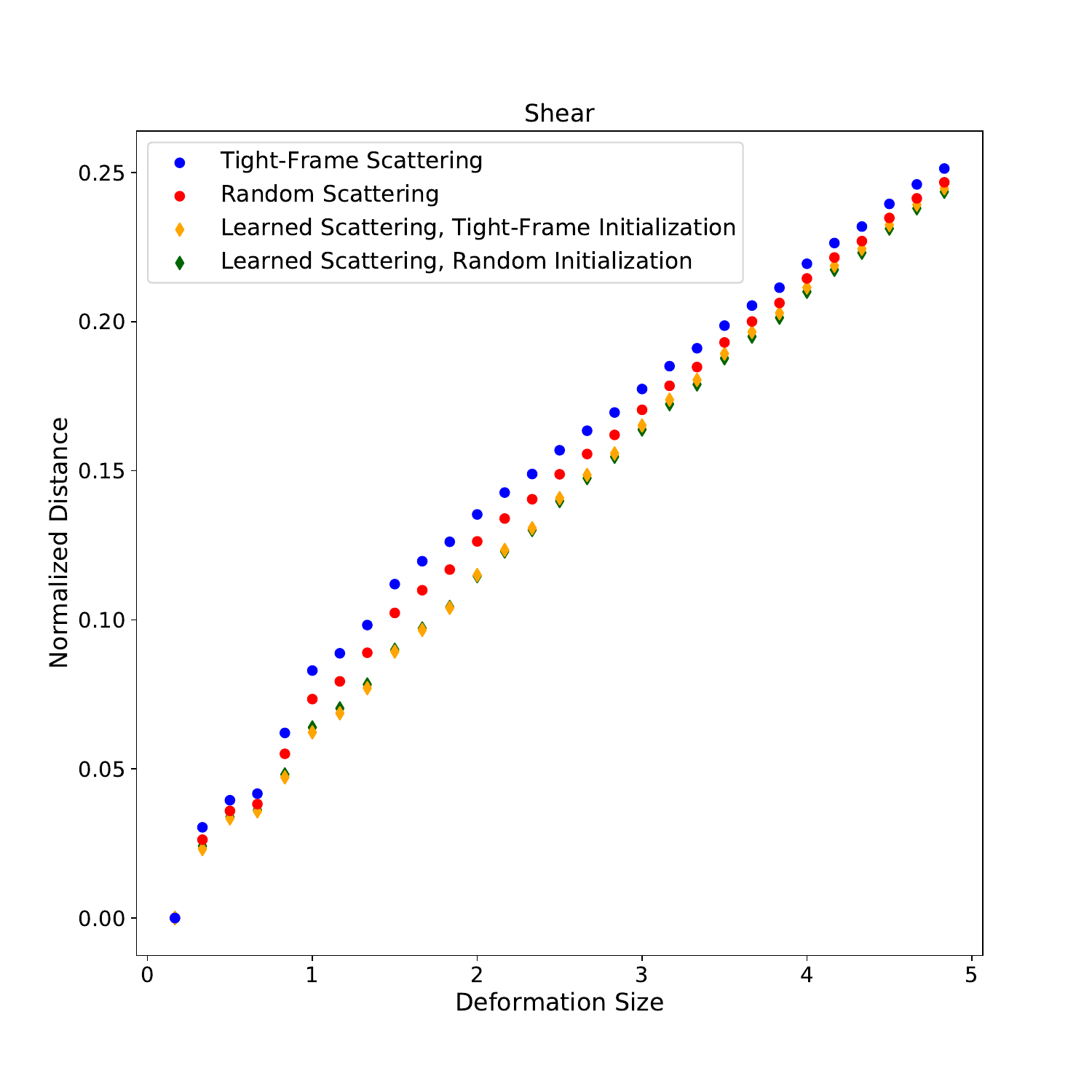}
    \includegraphics[width=0.27\linewidth]{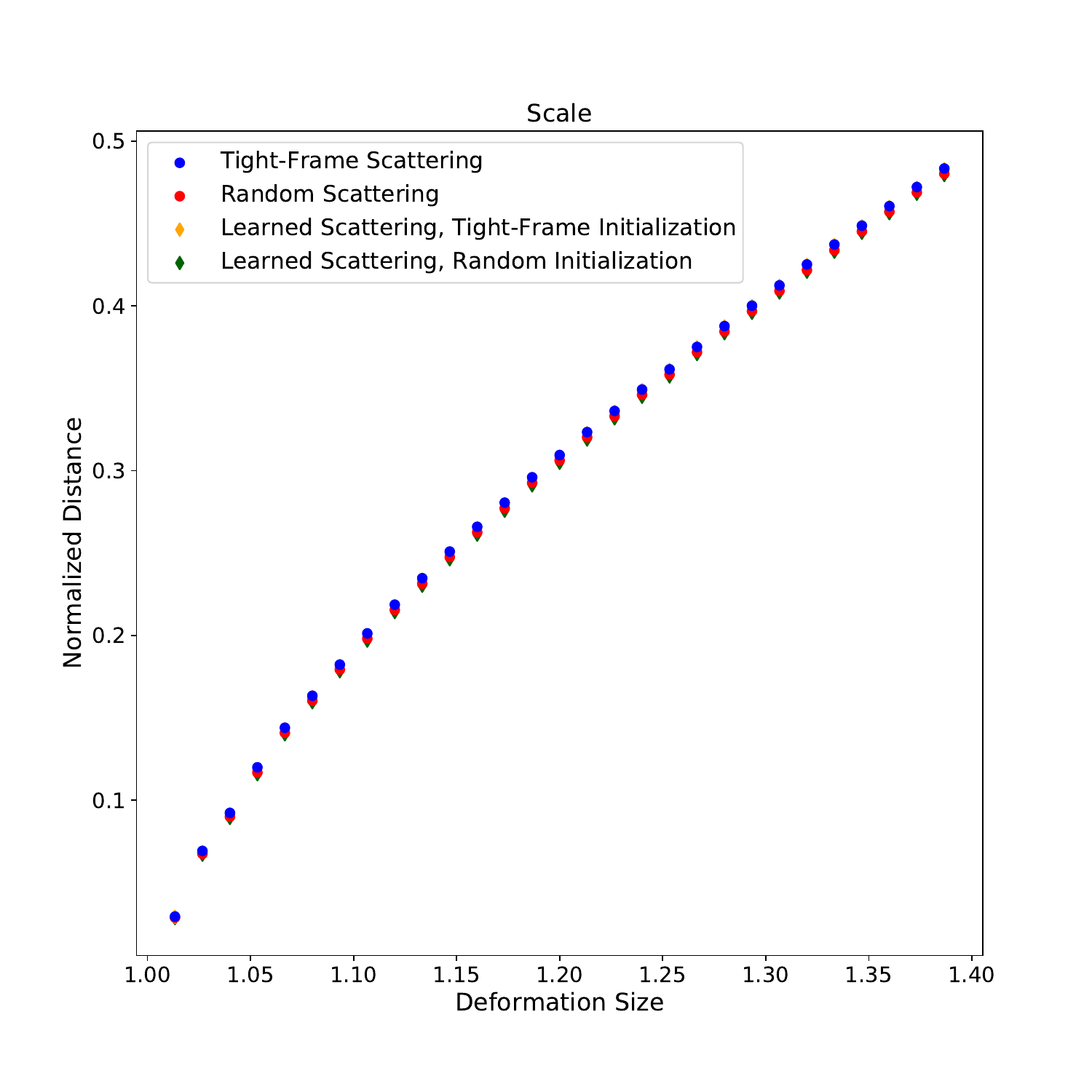}
    \vspace{-15pt}
    \caption{\textbf{Normalized distances between scattering representations of an image and its deformation.} Our parametric scattering transform shares similar stability to deformations as the scattering transform.   }% A comparision  Scattering representations of an image $S(x)$ and its transformation $S(\tilde{x})$   Normalized  of an X-ray image $S(x)$ and it's transformation $S(\tilde{x})$.
    \label{fig:defo}
\vspace{-12pt}
\end{figure*}
We first compare dataset-specific Morlet wavelet parameterizations and evaluate their similarities to a tight frame. Specifically, we train our parametric scattering networks using the canonical Morlet wavelet formulation with a linear classification layer and qualitatively compare the similarities of the learned filter bank to the tight-frame initialization. To facilitate quantitative comparison, we use a distance metric for comparing the sets of Morlet wavelet filters and Morlet wavelet filterbanks (i.e., scattering network instantiations), allowing us to measure deviations from the tight-frame initialization. 

We evaluate distances between two individual Morlet wavelets as $\Upsilon(M_1,M_2) = \left\| (\sigma_1,\xi_1,\gamma_1)^T - (\sigma_2,\xi_2,\gamma_2)^T  \right\|_2 + \textrm{arcdist}(\theta_1,\theta_2)
$ where $M_i=(\sigma_i,\xi_i,\gamma_i,\theta_i)^T$ denotes the parameterization of the Morlet wavelet. We use the arc distance on the unit circle to compare values of theta. Since the set of learned scattering filters does not have a canonical order, to compare a learned scattering network to the tight frame scattering network, we use a matching algorithm to match one set of filters to another. Specifically, we first compute $\Upsilon$ between all combinations of filter pairs from both networks, then use a minimum cost bipartite matching algorithm \cite{kuhn1955} to find the minimal distance match between the two sets of filters. The final distance we use as a notion of similarity between two scattering networks is the sum of $\Upsilon$ for all matched pairs in the bipartite graph. Henceforth, we will refer to this distance as the \textit{filterbank distance}. 

%Then, we compute the minimum cost bipartite matc

%To measure the distance between $\sigma,\xi,$ and $\gamma$ of two different initializations, computing the L$^2$ norm of their difference is reasonable. However, taking the L$^2$ norm of the difference between two $\theta$'s could lead to large distances between very similar filters. Instead, we compute the arc distance on the unit circle between both values of $\theta$.

%While the L$^2$ norm is a reasonable measure of distance for other parameters, taking the L$^2$ norm of the difference between two $\theta$'s could lead to large distances between very similar filters. Instead, we compute the arc distance on the unit circle between both values of $\theta$.
%Since the set of learnt scattering filters does not have a canonical order we propose to compare a learnt scattering network to the tight frame using a minimum cost bipartite matching algorithm \cite{kuhn1955} with $\Upsilon$ as the cost.  o compute the distance between the parameterizations of two scattering networks, we compute the Morlet wavelet distance between all possible combinations of their filters. 
%Then, we compute the minimum cost bipartite match \cite{kuhn1955} between these filters and use this value as our distance metric. 

The graph in Figure~\ref{fig:graphlwp} leverages the \textit{filterbank distance} to show the evolution of scattering networks initialized from a tight frame and trained on different datasets. Each network is trained on 1188 samples of its respective dataset (the standard size for KTH-TIPS2). All filters deviate quickly from a tight frame, but KTH-TIPS2's keep changing the longest and ultimately deviate the most. We also observe that filters initialized with the random initialization of Sec.~\ref{sec:parametric_scattering_networks} become more similar to our tight-frame initialization during the course of training (see Figure~\ref{fig:graphlwp_random} in Appendix~\ref{appendix:random-init}). 

On the left-hand side of Figure \ref{fig:graphlwp}, we visualize the dataset-specific scattering network parameterizations in Fourier space. White contours are drawn around each Morlet wavelet for clarity. The top black border corresponds to tight-frame initialization at J=2, shown for comparison to CIFAR-10 in blue (also J=2). The bottom black border corresponds to tight-frame initialization at J=4, shown for comparison to COVIDX-CRX2 red and KTH-TIPS2 yellow (both J=4).

The filters optimized on the KTH-TIPS2 texture dataset (yellow) become less orientation-selective (wider in Fourier space) than the tight-frame initialization, with filters at J=0 becoming the least orientation-selective of the whole filter bank. We note that the filters at spatial scales J= 2 and 3 seem to change the most from a tight frame as illustrated in the appendix (Fig. \ref{fig:kthsorted}). In contrast, the filters optimized on COVIDx-CRX2 become more orientation-selective in general i.e., thinner in Fourier space, while changing the most at spatial scale J=0 as shown in the appendix (Fig. \ref{fig:covidsorted}). The filters optimized on CIFAR-10 mirror those optimized on COVIDx-CRX2, also becoming more orientation-selective than their tight-frame counterparts. We suspect that this is due to a reliance on edges for object classification datasets, which seem to require more orientation-selective filters. On the other hand, the morlet wavelets optimized for texture classification seem to discard some edge information in favor of less orientation-specific filters. Each dataset-specific parameterization seems to discard unneeded information from the tight-frame initialization in favor of accentuating problem-specific attributes. In Sec.~\ref{sec:smalldata}, we demonstrate these learned filters are not only interpretable but improve task performance, suggesting the tight frame is not optimal for many problems of interest. Nonetheless, a tight-frame does constitute a good starting point for learning. Indeed, the dataset-specific parameterizations for COVIDX-CRX2 and KTH-TIPS2 are, visually, very different, yet they move similar filterbank distances from the tight-frame initialization (see fig.\ref{fig:graphlwp}), which are small relative to the distances observed for randomly initialized and trained models. 
\begin{table*}[t] 
    \centering
    \caption{CIFAR-10 mean accuracy and std.\ error over 10 seeds, with $J=2$ and multiple training sample sizes. Learnable scattering with TF initialization improves performance for all architectures, while randomly initialized scattering requires more training data to reach similar performance.} 
    \label{table:cifarresults}
    \fontsize{8}{8.5}\selectfont 
    \begin{tabularx}{0.75\linewidth}{lllllll} 
          \hline
        Arch. &Init. &Parameterization & 100 samples &500 samples & 1000 samples & All \\
        \hline
        \\[-2mm]
        LS+LL$\dagger$&TF&Canonical&$37.84\pm0.57$ & $\mathbf{52.68}\pm0.31$ &$\mathbf{57.43}\pm0.17$&$\mathbf{69.57}$ \\ 
        LS+LL$\dagger$&TF&Equivariant& $\mathbf{39.69}\pm0.56$ & $51.98\pm0.25$ &$57.01\pm0.16$ & $66.65$\\
        LS+LL&TF&Pixel-Wise&$32.30 \pm 0.69$ & $47.14 \pm 0.91 $ &$ 51.87 \pm 0.34$& $64.53$\\
        S +LL&TF&-&$36.01\pm0.55$ & $48.12\pm0.25$ &$53.25\pm 0.24$&$65.58$ \\
        LS+LL$\dagger$&Rand&Canonical& $34.81\pm0.60$ & $49.6\pm0.39$ &$55.72\pm0.39$&$69.39$ \\ 
        LS+LL$\dagger$&Rand&Equivariant&$34.67\pm0.73$ & $46.59\pm0.60$ &$52.95\pm0.36$&$65.64$ \\ 
        LS+LL &Rand&Pixel-Wise&$29.44 \pm 0.41$ & $42.14 \pm 0.27$ &$47.44 \pm 0.43$&$62.72$ \\
        S +LL&Rand&-& $29.77\pm0.47$ & $41.85\pm0.41$ &$46.3\pm0.37$&$57.72$ \\ % &0.155530 M   \\ 
        \hline
        \\[-2mm]
         LS+WRN$\dagger$&TF &Canonical&$\mathbf{43.60}\pm0.87$ & $\mathbf{63.13}\pm0.29$ &$70.14\pm0.26$&$93.61$ \\ 
         LS+WRN$\dagger$&TF &Equivariant&$ 39.86\pm1.59$ & $62.85\pm0.32$ &$69.52\pm 0.23$&$ 92.57 $  \\  
         LS+WRN&TF &Pixel-Wise&$39.20\pm0.80$ & $54.14\pm0.68$ &$57.59\pm0.48 $ &$ 92.97$ \\
        S +WRN&TF &-&$43.16\pm0.78$ & $61.66\pm0.32$ &$68.16\pm0.27$ &$92.27$ \\ 
       LS+WRN$\dagger$& Rand &Canonical& $41.42\pm0.65$ & $59.84\pm0.40$ &$67.40\pm0.28$&$93.36$ \\
        LS+WRN$\dagger$&Rand &Equivariant&$ 40.84\pm1.02$ & $60.81\pm0.40$ &$68.62\pm0.31$&$ 92.53$  \\
        LS+WRN& Rand &Pixel-Wise&$31.49\pm0.63$ & $45.85\pm0.43$ &$50.72\pm0.28 $&$ 91.86 $  \\
        S +WRN&Rand &-&$32.08\pm0.46$ & $46.84\pm0.21$ &$52.76\pm0.33$&$85.35$ \\
        WRN-16&-&-& $38.78\pm 0.72$ & $62.97\pm 0.41$ &$\mathbf{71.37}\pm0.31$& $\mathbf{96.84}$  \\
       ResNet-50&-&-& $33.17 \pm0.92$ & $ 52.13\pm0.74 $ &$64.42\pm0.40 $&$ 91.23$   \\ % & 22.320272 M \\
        \hline
    \end{tabularx}
\begin{flushleft}
\scriptsize
\vspace{-5pt}
\hspace{65pt}$\#$ params : 156k for S+LL; 155k for LS+LL; 22.6M for S+WRN;  22.6M for LS+WRN; 22.3M for WRN; and 22.5M for ResNet\\
\hspace{65pt}$\dagger$: ours;\hspace{5pt}  TF: Tight-Frame;\hspace{5pt}LS: Learnable Scattering;\hspace{5pt}S: Scattering;\hspace{5pt} Rand: Random\\
\end{flushleft}
\vspace{-20pt}
% \\[-5mm]
\end{table*}

\subsection{Robustness to Deformation}
\label{section:deformation}
In \cite{mallat2012group}, it is shown that the scattering transform is stable to small deformations of the form $x(u-\tau(u))$ where $x(u)$ is a signal and $\tau$ a diffeomorphism. Given the substantial changes to the filter composition in the learning process, we ask now whether these seem to significantly deviate from the stability result obtained from the carefully handcrafted construction proposed in~\cite{mallat2012group} and extensively used in previous work e.g.,~\cite{bruna2013invariant,eickenberg2018solid}. To evaluate the robustness of our parametric scattering networks to different geometric distortions, we apply several tractable deformations to a chest X-ray image $x$ with varying deformation strength and encode all images with different (learned and fixed) scattering networks. The learned ones are trained using the Morlet canonical wavelet formulation with a linear classification layer. The transformed image is denoted by $\tilde{x}$. For each of the different deformation strengths, we plot the Euclidean distance between the scattering feature constructed from the original image $S(x)$ and the scattering feature constructed from the transformed image $S(\tilde{x})$. We then normalize the obtained distance by $S(x)$ to measure the relative deviation in scattering coefficients (handcrafted or learned). Figure \ref{fig:defo} demonstrates representative results for a small rotation, shear and scale on images from the COVIDx datasets, while additional deformations are shown in Appendix \ref{appendix:deformations}. We observe that the substantial change in the filter construction retains the scattering robustness properties for these simple deformations, thus indicating that the use of learned filters (instead of designed ones) does not necessarily detract from the stability of the resulting transform.

\begin{table*}[t] 
    \centering
    \vspace{-10pt}
    \caption{COVIDx CRX-2 and KTH-TIPS2 mean accuracy \& std.\ error with $J=4$ over 10 seeds and 16 seeds respectively. (COVIDx CRX-2) TF-initialized learnable scattering network performs better than models that do not incorporate scattering priors. (KTH-TIPS2) Similarly, the WRN-16-8 and ResNet-50 perform extremely poorly relative to hybrid models trained on KTH-TIPS2.}
    \label{table:covidresults}
      \fontsize{8}{8,5}\selectfont 
   % \adjustbox{width=390pt}{
    \begin{tabularx}{0.8\linewidth}{llllll|ll} 
             \hline
             % \\%[-2mm]
            Arch.&Init. & Parameterization&C-100 samples & C-500 samples &C-1000 samples & KTH-1188 samples  \\
            \hline
            LS+LL$\dagger$&TF &Canonical& $82.30\pm1.78$ &$\mathbf{88.50}\pm0.71$ & $\mathbf{89.90}\pm0.40$ &$ 66.09\pm1.05 $  \\ 
            LS+LL$\dagger$&TF &Equivariant&$\mathbf{83.06}\pm1.53$ & $87.56\pm0.94$ &$89.15\pm0.60$ &$ \mathbf{66.41}\pm1.24$ \\
            S +LL&TF&-& $81.08\pm1.88$&$87.20\pm0.77$ &  $89.23\pm0.69$ &$66.17\pm1.10$\\
            LS+LL$\dagger$&Rand  &Canonical& $76.85\pm1.50$&$86.45\pm0.95$ & $89.70\pm0.65$& $65.79\pm 0.85$ \\
            LS+LL$\dagger$&Rand  &Equivariant&$76.73\pm1.57$ & $85.64\pm1.38$ &$87.98\pm0.55$ & $65.31\pm1.42$ \\
            S +LL&Rand &-& $76.08\pm1.56$&$84.13\pm0.91$ &  $86.80\pm0.41$ &$ 61.37\pm 0.82$\\[1mm]
            \hline
            % \\%[-2mm]
            LS+WRN$\dagger$&TF &Canonical& $81.20\pm1.73$&$90.50\pm0.70$ & $93.68\pm0.35$&$\mathbf{69.23}\pm0.67$\\
            LS+WRN$\dagger$&TF &Equivariant&$\mathbf{81.86}\pm2.07$ & $\mathbf{91.56}\pm0.52$&$\mathbf{93.97}\pm0.34$&$68.55\pm0.80$\\
            S +WRN&TF &-&$80.85\pm1.85$&$89.05\pm0.59$ & $91.90\pm0.54$ &$68.84\pm0.71$ \\
            LS+WRN$\dagger$&Rand &Canonical&$80.95\pm1.54$&$88.08\pm0.70$ & $91.65\pm0.55$&$68.30\pm0.47$ \\
            LS+WRN$\dagger$&Rand &Equivariant&$80.12\pm1.76$ & $87.44\pm1.17$&$91.40\pm0.67$&$ 67.50\pm0.72$\\

            S +WRN&Rand &-& $80.63\pm1.73$&$86.68\pm0.59$ & $90.60\pm0.50$ & $66.29\pm0.36$\\
             WRN-16&-&- &80.50$\pm 1.15$ &  $85.95\pm2.04$ & $88.82\pm1.64$ & $51.24\pm1.37$\\
             ResNet-50&-&-& $74.04 \pm1.35$ & $86.45 \pm 0.51$ &$90.86\pm 0.57$ &$44.95\pm 0.65$  \\
            \hline
        \end{tabularx}
\begin{flushleft}
\scriptsize
\vspace{-5pt}
\hspace{55pt}C: COVIDx CRX-2\hspace{5pt} $\#$ params : 493K for LS/S+LL; 23.7M for LS/S+WRN; 22.3M for WRN;23.5M for ResNet\\
\hspace{55pt}KTH: KTH-TIPS2\hspace{5pt} $\#$ params : 883K for LS/S+LL; 23.8M for LS/S+WRN; 22.3M for WRN; 23.5M for ResNet\\
\hspace{55pt}$\dagger$: ours; \hspace{5pt} TF: Tight-Frame;\hspace{5pt}LS: Learnable Scattering;\hspace{5pt}S: Scattering;\hspace{5pt} Rand: Random
\end{flushleft}
% \\[-5mm]
\vspace{-20pt}
\end{table*}

\subsection{Small Data Regime}
 \label{sec:smalldata}
 We evaluate the parametric scattering network in limited labeled data settings. Following the evaluation protocol from \cite{oyallon2018replearning}, we subsample each dataset at various sample sizes to showcase the performance of scattering-based architectures in the small data regime.  In our experiments, we train on a small random subset of the training data but always test on the entire test set as done in~\cite{oyallon2018replearning}. To obtain comparable and reproducible results, we control for deterministic GPU behavior and assure that each model is initialized the same way for the same seed. Furthermore, we use the same set of seeds for models evaluated on the same number of samples. For instance, the TF learnable hybrid with a linear model would be evaluated on the same ten seeds as the fixed tight-frame hybrid with a linear model when trained on 100 samples of CIFAR-10. Some fluctuation is inevitable when subsampling datasets. Hence all our figures include averages and standard error calculated over different seeds. 
 \vspace{-10pt}
\paragraph{CIFAR-10} \hspace{-8pt}
consists of 60,000 images from ten classes. The train set contains 50,000 class-balanced samples, while the test set contains the remaining images. Table~\ref{table:cifarresults} reports the evaluation of our learnable scattering approach on CIFAR-10 with training sample sizes of 100, 500, 1K, and 50K. The training set is augmented with horizontal flipping, random cropping, and pre-specified autoaugment~\cite{autoaugment} for CIFAR-10. We used autoaugment \cite{autoaugment} to showcase the best possible small-sample results and ablate this component in Appendix \ref{appendic:noautoaugment}. We use a spatial scale of $J=2$ in the scattering transforms. 

As shown in Table \ref{table:cifarresults}, the scattering networks with wavelets optimized pixel-wise perform the worst in the small-data regime. It shows that with limited labeled samples, there is not enough data and too many learnable parameters to learn effectively the pixels of the wavelets. Adding more constraints (i.e., constraining the wavelets to be Morlet) is beneficial in this setting.  We also observe that the Morlet canonical parameterization yields a similar performance to the Morlet equivariant parameterization (i.e., most standard errors overlap). Thus, adding even more constraints, by reducing the number of learnable parameters in the parametric scattering transform, does not degrade the performance in the small-data regime. %When the full training set is used, we observe that the canonical parameterization yields better accuracy than the equivariant parameterization, showing that when the number of labeled data is not limited, adding more constraints can decrease the performance.
We observe that randomly initialized learnable with canonical parameterization only achieves similar performance to TF learnable canonical when trained on the whole dataset. These results suggest the TF initialization, derived from rigorous signal processing principles, is empirically beneficial as a starting point in the very few sample regime but can be improved upon by learning.
\begin{table*}
    \caption{Scattering and learned unsupervised scattering features evaluated by training a linear classifier on CIFAR-10. We observe the unsupervised learned scattering improves the representation.\vspace{-10pt}}
    \centering
    \small
    \label{table:unsup}
      \fontsize{8}{8.5}\selectfont 
    \begin{tabular}{lllll}
             \hline
          Method &100 samples & 500 samples & 1000 samples  & All\\
            \hline
           Scattering (Fixed) & $36.01\pm0.55$ & $48.12\pm0.25$ &$53.25\pm 0.24$&$65.58\pm0.04$ \\ 
           Unsupervised Learnt Scattering &$\textbf{38.05} \pm 0.45 $ & $ 	\textbf{52.92} \pm 0.28	$ &$\textbf{57.76} \pm 0.25 $&$ \textbf{68.47} \pm 0.04$ \\ 
            \hline
         %  Unsup R Param Scat. &$36.54 \pm 0.5 $ & $ 	50.61 \pm 0.29	 $ &$ 66.82 \pm 0.04  $&$ $ \\ 
%cocuou 
\end{tabular}
\vspace{-8pt}
\end{table*}
Among the linear models, our TF-initialized learnable scattering networks (i.e., Morlet canonical and equivariant) significantly outperform all others in few sample settings. This demonstrates that learnable scattering networks obtain a more linearly separable representation than their fixed counterparts, perhaps by building greater dataset-specific intra-class invariance. 
%Interestingly, when comparing TF learnable Morlet canonical to fixed, we observe that the relative performance gain of learning increases from 100 samples ($1.83$) to 500 samples ($4.56$), yet it remains relatively constant for 500, 1K, and 50K sample settings, indicating that perhaps 500 samples are sufficient to tune a scattering representation when starting from TF initialization. 
%When comparing TF learnable Morlet canonical equivariant to fixed, we observe a different behavior. The relative performance gain of learning remains relatively constant from 100 samples ($3.68$) to 1K samples ($3.76$) and then drops at 50K ($1.07$). At 50K, the extra constraints of the equivariant wavelets harm the performance. 
Figure~\ref{fig:filters-cifar-real-before-kymatio} shows the real part of the canonical wavelet filters before and after optimization on the entire training set. In Appendix \ref{appendix:equivariant}, we visualize canonical equivariant wavelet filters. 

Among the WRN hybrid models, the TF-initialized canonical learnable scattering performs best.  Canonical TF learnable still improves over TF fixed when paired with a WRN, indicating some loss of information in the fixed scattering representation is mitigated by data-driven tuning or optimization. Finally, our approach outperforms the fully trained ResNet-50 and outperforms the WRN-16-8 on 100 and 500 training samples, demonstrating the effectiveness of the scattering prior in the small data regime. However, the WRN-16-8 outperforms our model on 1,000 samples and 50,000 samples.

 \vspace{-10pt}
\paragraph{COVIDx CRX-2} \hspace{-8pt}
\label{section-covid} is a two-class (positive and negative) dataset of chest X-Ray images of COVID-19 patients~\cite{Wang2020}. The train set contains 15,951 unbalanced images, while the test set contains $200$ positive and $200$ negative images. The spatial scale of the scattering transform is set to $J=4$. Table \ref{table:covidresults} reports our evaluation on sample sizes of 100, 500, and 1K images. We use the same protocol as for CIFAR-10. Morlet canonical parameterization yields similar performance to the Morlet equivariant parameterization (i.e., most standard errors overlap), as also observed with CIFAR-10.

When the scattering networks are postpended with a linear layer, TF-initialized learnable (i.e., Morlet canonical and equivariant) performs better than TF fixed, showing the viability of our approach on real-world data. We observe that randomly initialized learnable yields lower performance than TF learnable on 100 and 500 samples. On 1K, it achieves similar performance, demonstrating that random initialization can achieved comparable performance to TF with enough data.  WRN-16-8 performs worse than TF-initialized learnable followed with a linear layer. When combined with a CNN,  TF-initialized learnable also performs better than TF fixed and outperforms WRN-16-8 and ResNet-50.
\vspace{-10pt}
\paragraph{KTH-TIPS2} \hspace{-8pt} \hspace{-8pt} contains 4,752 images from 11 material classes. The images captured the material at scales.  Each class is divided into four \emph{samples} (108 images each) of different scales.  Using the standard protocol, we train the model on one \emph{sample} ($11*108$ images), while the rest are used for testing~\cite{song2017locally}. In total, each training set contains 1,188 images. Table \ref{table:covidresults} reports the classification accuracies. With TF initialization and a linear layer, we observe that the performance is similar for the different architectures. The performance of randomly initialized learnable is also similar to TF. The fixed and randomly initialized model perform the worst, showing that even poorly initialized filters can effectively be optimized. Altogether, these results further corroborate our previous findings, notably that TF initialization acts as a good prior for scattering networks. Out of all the WRN hybrid models, the TF-initialized learnable model using canonical parameterization achieves the highest average accuracy. We note that while WRN increases the performance compared to the linear layer, it also significantly increases the total number of parameters, therefore exhibiting a tradeoff between performance and model complexity. The WRN-16-8 and ResNet-50 perform extremely poorly relative to hybrid models, showing the effectiveness of the scattering priors for texture discrimination.

\subsection{Unsupervised Learning of Parameters}
\label{sec:unsup}
We have studied the adaptation of the wavelet parameters towards a supervised task. We now perform a preliminary investigation to determine if the scattering representation can be improved in a purely unsupervised manner. We consider the recently popularized SimCLR framework \cite{chen2020simple}, which encourages representations from two data augmentations of the same input to lie close together. We learn scattering network parameters with the canonical Morlet parameterization on CIFAR-10 using this unsupervised objective function and subsequently evaluate the discriminativeness of the features under a standard linear evaluation experiment on the full CIFAR-10 dataset and in the small data regimes comparing them to the standard scattering transform. The results are shown in Table~\ref{table:unsup}. We observe the unsupervised learning of filter parameters can improve the scattering representation under standard unsupervised learning evaluation protocols. % outperforms the scattering network across all small sample and full data settings. 

\subsection{Computational and memory complexity}
\label{sec:computation}
%While our empirical evaluation demonstrates the strong performance of parametric scattering in the small sample regime, computational and memory complexity are also important considerations. 
The computational complexity of scattering networks and parametric scattering networks is directly related to the FFT (Fast Fourier transform), $O(N{\cdot}\log(N))$ for an image of size (N $\times$ N). %The Kymatio package~\cite{andreux2020kymatio} offers an efficient GPU implementation and is used in our implementation. 
In practice, the computational and memory complexity of our parametric scattering networks varies due to a number of factors.%: hyperparameters, hybrid architecture, image size, and more. 
 To summarize these factors, we compare runtime (higher is faster), memory, and parameter count per architecture and image size in Table~\ref{table:complexity}. The models were trained using an NVIDIA Tesla T4 GPU. We observe that fixed scattering is two to three times faster than learned scattering for all image sizes and hybrid models. In contrast, WRN-16-8 is faster than LS+WRN at image size $32^2$, but slower for larger images. This is due to the scattering transform's substantial spatial dimension reduction, which leads to speed and memory benefit versus regular CNNs~\cite{mallat2012group}. While gradient computation of Morlet parameters adds compute overhead, learned scattering is still efficient with much fewer parameters than CNNs.

% Strictly speaking, the computational complexity of scattering networks and parametric scattering networks is directly related to the FFT (Fast Fourier transform), yielding $O(N{\cdot}\log(N))$ for an image of size (N $\times$ N). However, we leverate the The Kymatio package~\cite{andreux2020kymatio} offers an efficient GPU implementation and is used to conduct the experiments. Further, the scattering networks followed by CNNs for larger images benefit from substantial spatial dimension reduction by scattering, which leads to speed and memory benefit versus regular CNNs~\cite{mallat2012group}.
% Table~\ref{table:complexity} compares runtime (higher is faster), memory, and parameter count per architecture and image size. The models were trained using an NVIDIA Tesla T4 GPU. While gradient computation of Morlet parameters adds compute overhead, learned scattering is still efficient with much fewer parameters than CNNs.

% Computational and memory efficiency of LS+L is worse than S+L but still highly competitive vs CNNs esp.\ with larger images. Further imp-rovements in train. time/mem.\ effic- iency are possible via custom CUDA kernels \& architecture optimization but out of scope in this work.
% of architectures and images sizes in the paper. 
%GPU usage is measured batches of size 128  .

\begin{table}
    \caption{Comparison of training runtime, inference runtime, GPU memory, and parameter count per architecture and image size.  \\ \vspace{-20pt}}
    \centering
    \small
    \label{table:complexity}
      \fontsize{8}{8}\selectfont 
    {\begin{tabular}{lcccccl}
    \hline 
    Architecture &\parbox{10pt}{Img.\\ size} & \parbox{10pt}{Train\\($\frac{\text{imgs}}{\text{sec}}$)}	& \parbox{10pt}{Infer.\\($\frac{\text{imgs}}{\text{sec}}$)} & \parbox{10pt}{\scriptsize{}GPU Mem.\\(GB)} & \parbox{25pt}{\#Params\\(Million)} & \\
    \hline 
    LS+L  & $32^2$ & $264$ &	$542$ & $	0.3	$	&	$0.2$ & \multirow{5}{*}{\hspace{-14pt}\rotatebox[origin=c]{-90}{$\overbrace{\hspace{44pt}}^{\text{CIFAR}}$}}\\
    S+L & $32^2$ &	$650$ &	$656$ & $	0.1	$	& $0.2$ & \\
    LS+WRN  & $32^2$& $232$ & $430$	& $	1.6	$	& $22.6$ & \\
    S+WRN  & $32^2$ & $498$ & $588$ & 
    $	1.4	$	& $22.6$ & \\
    WRN-16-8*	& $32^2$ &	$510$ &	$1695$ & 
    $	4.2	$	& $11.0$ & \\\hline
    LS+L  & $128^2$ &	$42$ &	$102$ & 
    $	5.4	$	&	$0.8$ & \multirow{5}{*}{\hspace{-14pt}\rotatebox[origin=c]{-90}{$\overbrace{\hspace{43pt}}^{\text{KTH}}$}}\\
    S+L & $128^2$ &	$123$ &	$123$ & 
    $	0.5	$	&	$0.8$ & \\
    LS+WRN & $128^2$ & $42$ &	$101$ &  $	8.5	$	&	$23.8$ & \\
    S+WRN & $128^2$ &	$115$ &	$120$ & 
    $	3.3	$	&	$23.8$ & \\
    WRN-16-8* & $128^2$	& $31$ & $90$ & 
    $	61.1	$	&	$11.0$ & \\\hline
    LS+L & $224^2$ & $24$	& $72$ & 
    $	13.7	$	& $0.5$ & \multirow{5}{*}{\hspace{-14pt}\rotatebox[origin=c]{-90}{$\overbrace{\hspace{43pt}}^{\text{COVID}}$}}\\
    S+L & $224^2$ & $78$ &	$79$ &  $	1.3	$	& $0.5$ &\\
    LS+WRN & $224^2$ &	$22$ & $67$	& 
    $	16.1	$	& $23.7$ &\\
    S+WRN & $224^2$ & $58$	& $71$ & 
    $	2.9	$	& $23.7$ &\\
    WRN-16-8* & $224^2$ & $10$ & $36$ & $	49.6	$	& $11.0$ &\\
    \hline
    \end{tabular}}
    \begin{flushleft}
\scriptsize
\vspace{-7pt}
\hspace{18pt}$*$ from \cite{zagoruyko2016wide}\\
\end{flushleft}
\vspace{-18pt}
\end{table}

%Although below S+L,
% regular scattering, 
\section{Limitations}
There are two limitations to this study that could be addressed in future research. First, the current implementation is limited to two-dimensional data. In future work, the implementation could naturally be extended to one-dimensional and three-dimensional data. Second, for popular datasets, such as CIFAR-10, there are pre-trained models available.  In the study, to compare performance with our approach, we considered a fully learned WRN-16-8 and ResNet-50, but we did not consider pre-trained models. 
%On large datasets, our approach does not seem to perform as well as these pre-trained models.

\section{Conclusion}
\label{sec:conclusion}
% This work demonstrated the competitive results of adapting a small number of Morlet wavelet filter parameters in the scattering network. We illustrated that filters learned by parametric scattering can be interpreted in relation to the specific task (e.g., becoming thinner in object recognition tasks that require sensitivity to edges). We also empirically demonstrate that our parametric scattering transform shares similar stability to deformations as the traditional scattering transforms. Overall we find that the parametric scattering network provides state-of-the-art results for classification in the low-data regime when combined with a linear layer and as well in a hybrid CNN. These results go towards bridging the gap between the handcrafted filter design in scattering transforms, which provides tractable properties and supports low-parameter models, and the fully (parameterized) learned ones commonly used in CNN work, especially in computer vision and generally on 2D structured data. In particular, our results can lead to future work investigating the impact of downsampling on the representations learned by the parametric scattering network, as well as application to uncertainty estimation by leveraging the low parameter CNN in a Bayesian framework.

This work showcases the competitive results of adapting a small number of Morlet wavelet filter parameters in scattering convolutional networks \cite{bruna2013invariant}. We demonstrate that filters learned by parametric scattering can be interpreted in the context of specific tasks (e.g., becoming thinner in object classification tasks that require sensitivity to edges). We also empirically demonstrate that our parametric scattering transform shares similar deformation stability to the traditional scattering transform. Overall, we find that our hybrid parametric scattering architectures (with LL and WRN) achieve state-of-the-art classification results in the low-data regime. These results verge upon bridging the gap between the handcrafted filter design in traditional scattering transforms, which provides tractable properties and supports low-parameter models, and fully parameterized convolutional neural networks, which lack interpretability but are more flexible.

 In the future, our results can lead to work investigating the impact of downsampling on the representations learned by the parametric scattering network, as well as application to uncertainty estimation by leveraging the low parameter CNN in a Bayesian framework.

%-------------------------------------------------------------------------

%%%%%%%%% REFERENCES
{\small
\bibliographystyle{ieee_fullname}
\bibliography{egbib}
}
\clearpage
\appendix
\setcounter{page}{1}
\onecolumn
\section{Implementation Details}
\label{appendix-details}
This section describes the implementation details for each dataset.
\subsection{CIFAR-10}
\label{appendix:cifar}
CIFAR-10 consists of consists of 60,000 images of size $32 \times 32 \times 3$ from ten classes. The linear models were trained using a max learning rate of 0.06 for all parameters on 5K, 1K, 500, and 500 epochs for 100, 500, 1K, 50K samples, respectively. The hybrid WRN models were trained using a max learning rate of 0.1 on 3K, 2K, 1K, and 200 epochs for 100, 500, 1K, and 50K samples respectively. We use batch gradient descent except when the models are trained with 50K samples where we use mini-batch gradient descent of size 1024. On the entire training set, we also train the models on 10 seeds and, in all cases, the standard errors are always inferior to $0.3 $. All scattering networks use a spatial scale $J=2$.

\subsection{COVIDx CRX-2}
COVIDx CRX-2 is a two-class (positive and negative) dataset of $1024 \times 1024 \times 1$ chest X-Ray images of COVID-19 patients~\cite{Wang2020}. In our experiments, we always train on a class-balanced subset of the training set. We resize the images to $260 \times 260$ and train our network with random crops of $224 \times 224$ pixels. The only data augmentation we use is random horizontal flipping. All models were trained on 400 epochs using a max learning rate of 0.01. All hybrid models are trained with a mini-batch size of 128. All scattering networks use a spatial scale $J=4$.
\subsection{KTH-TIPS2}
\label{appendix:detail-kth}
We resize the images to $200 \times 200$ and train our network with random crops of $128 \times 128$ pixels. The training data is augmented with random horizontal flips and random rotations. All scattering networks use a spatial scale of 4. We set the maximum learning rate of the scattering parameters to 0.1 while it is set to 0.001 for all other parameters. All hybrid models are trained with a mini-batch size of 128. The hybrid linear models are trained for 250 epochs, while the hybrid WRN models are trained for 150 epochs. We evaluate each model, training it with four different seeds on each \emph{sample} of material, amounting to 16 total runs. All scattering networks use a spatial scale $J=4$.

\section{Wide Residual Network Hybrid Architecture}

\label{appendix-wrn}
In the experiments of Sec.~\ref{sec:results}, the scattering networks are combined with a WRN hybrid described in \cite{oyallon2017scaling}. We follow a similar architecture to the WRN hybrid used in \cite{oyallon2018replearning}. The description of the architecture used for the experiments is given in Table~\ref{tab:scat_wrn}. We use the same architecture for all the datasets. The architecture consists of a scattering network that greatly reduces the spatial resolution of the input followed by a WRN. The CIFAR-10 scattering stage yields output with $8\times8$ spatial resolution (scattering with J=2). Similarly, the KTH-TIPS2 data and COVIDx-CRX2 data give outputs with $16\times16$ and $8\times8$ spatial resolutions respectively. 
%However, in~\cite{oyallon2018replearning}, scattering networks are combined with a Wide Residual Network \cite{wideresnet}. We replicate the experiments of Sec.\ref{sec:results} with learnable scattering, but this time the scattering networks are followed by the WRN of depth 16 and width 8 used in ~\cite{oyallon2018replearning}. 

\begin{table}[H]
\setlength{\tabcolsep}{15pt}
\renewcommand{\arraystretch}{1.5}
    \centering
    \begin{tabular}{|c|c|}
    \hline
        \textbf{Stage} & \textbf{Description} \\\hline
        scattering &  Learned or Not Learned
        \\\hline
        conv1 & $3\times3$, \textsc{Conv Layer } $128\rightarrow256$ \\\hline
        conv2 & $\begin{bmatrix}3\times3, \textsc{Conv Layer } 256 \\ 3\times3, \textsc{Conv Layer } 256\end{bmatrix} \times 4$\\\hline
        conv3 & $\begin{bmatrix}3\times3, \textsc{Conv Layer }256 \\ 3\times3, \textsc{Conv Layer } 256\end{bmatrix} \times 4$\\\hline
        avg-pool &  Avg pooling to a size 1x1\\\hline
    \end{tabular}
    \caption{Description of the WRN hybrid architecture used for the experiments. Each convolutional layer represents a 2-dimensional convolution followed by a batch normalization and a ReLU non-linearity function.}
    \label{tab:scat_wrn}
\end{table}

\section{Backpropagation through the Parametric Scattering Network}
\label{sec:derivative}
% To provide such data-driven optimization of scattering parameters, 
We show here that it is possible to backpropagate through this construction. Namely, we verify the differentiability of this construction by explicitly computing the partial derivatives with respect to these parameters. First, the $\mathbb{R}$-linear derivative of the complex modulus $f(z) = |z|$ is $f'(z) = \frac{z}{|z|}$. Next, we show the differentiation of convolution with wavelets with respect to their parameters. For simplicity, we focus here on differentiation of the Gabor portion\footnote{It is not difficult to extend this derivation to Morlet wavelets, but the resulting expressions are rather cumbersome and left out for brevity.} of the filter construction from Eq.~\ref{eq:1}, written as:
\vspace{-5pt}
\begin{align*}
        \varphi(u) = &\exp(-\frac{1}{2\sigma^2}(u_1^2(\cos^2(\theta)+\sin^2(\theta) \gamma^2)\\&+u_2^2(\cos^2(\theta) \gamma^2 + \sin^2(\theta)) \\& + 2\cos(\theta) \sin(\theta) u_1 u_2 (1-\gamma^2)) \\& + i \xi (\cos(\theta) u_1 + \sin(\theta) u_2)). \tag{4}
\end{align*}
Its derivatives with respect to the parameters are
\vspace{-5pt}
\begin{align*}
    \frac{\partial \varphi}{\partial \theta} (u) =& \; \frac{1}{\sigma^2}(u_2 \cos\theta - u_1 \sin\theta)(i\xi \sigma^2 + u_1 (\gamma^2-1)\cos\theta \\& + u_2(\gamma^2-1)\sin\theta)\varphi(u);\tag{5}\\
    \frac{\partial \varphi}{\partial \sigma} (u) =& \; \frac{1}{\sigma^3} (u_1^2(\cos^2\theta+ \gamma^2 \sin^2\theta )+u_2^2(\gamma^2 \cos^2\theta +\sin^2\theta)  \\&+ 2 u_1 u_2 \cos\theta \sin\theta (1-\gamma^2)) \varphi(u); \tag{6} \\
    \frac{\partial \varphi}{\partial \xi} (u) =& \; i(u_1 \cos\theta + u_2 \sin\theta) \varphi(u); \tag{7} \text{ and} \\
    \frac{\partial \varphi}{\partial \gamma} (u) =& -\frac{1}{\sigma^2}(u_1^2 \gamma \sin^2\theta  + u_2^2 \gamma \cos^2\theta  \\&- 2 u_1 u_2 \gamma \cos\theta \sin\theta ) \varphi(u). \tag{8}
\end{align*}
Finally, the derivative of the convolution with such filters is given by $\frac{\partial}{\partial \zeta}(f*\varphi)(t) = \int f(t-u) \frac{\partial \varphi}{\partial \zeta} (u) du$ where $\zeta$ is any of the filter parameter from Table~\ref{tab:params}. It is easy to verify that these derivations can be chained together to propagate through the scattering cascades defined in Sec.~\ref{sec:scatteringnetworks}.  We can now learn these jointly with other parameters in an end-to-end differentiable architecture.

\section{Equivariant Filters}
\label{appendix:equivariant}
\begin{figure}[!ht]
    \centering
    \includegraphics[width=0.6\linewidth]{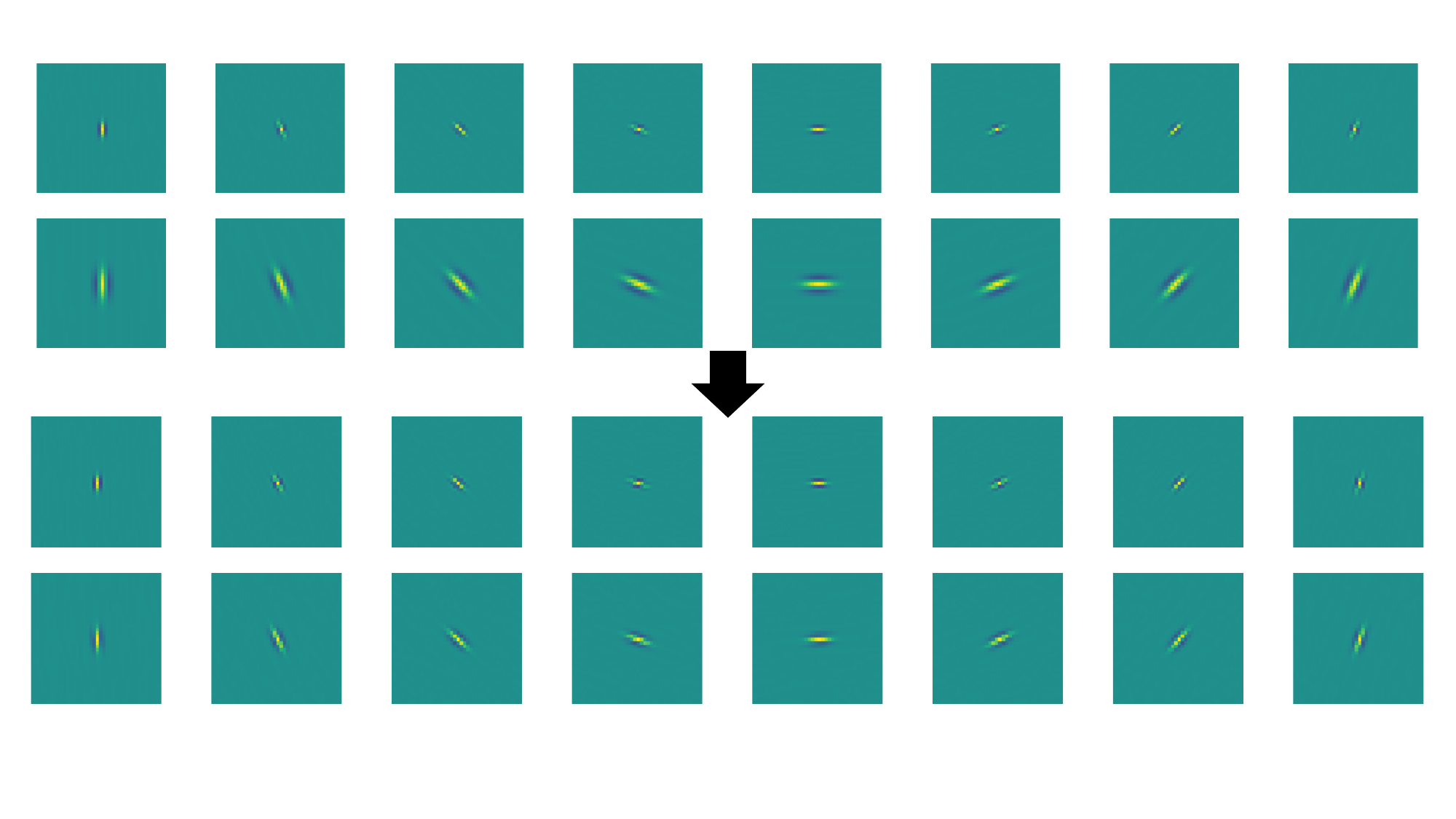}
        \vspace{-20pt}
    \caption{\textbf{Example of equivariant filters initialized using tight frame construction}. (Top) Real part of wavelet filters before optimization. (Bottom) Real part of wavelet filters after optimization.}
    \label{fig:equivarianttf}

\end{figure}

\begin{figure}[!ht]
    \centering
    \includegraphics[width=0.6\linewidth, trim={0 0 0 2cm}]{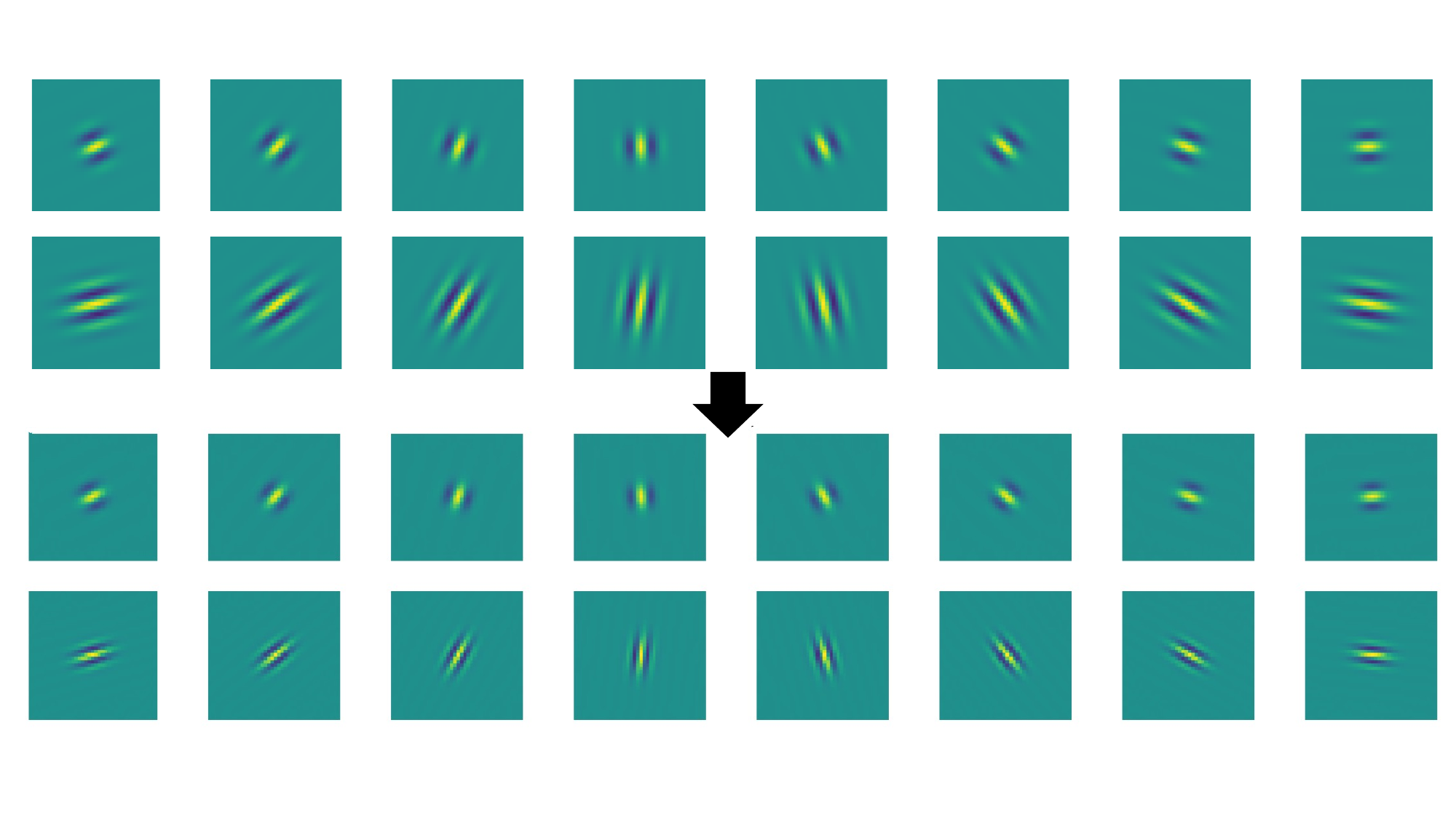}
    \vspace{-20pt}
    \caption{\textbf{Example of equivariant filters initialized randomly}. (Top) Real part of wavelet filters before optimization.  (Bottom) Real part of wavelet filters after optimization.}
    \label{fig:equivariantrd}
\vspace{-10pt}
\end{figure}

We observe that, in some cases, using equivariant filters yields better accuracy, as shown in Table \ref{table:cifarresults}.  Figure \ref{fig:equivarianttf} illustrates equivariant filters initialized using tight frame construction before and after optimization. The scattering network is combined with a linear layer and trained on 500 training samples of CIFAR-10. The spatial scale is set to $J=2$. Similarly, Figure \ref{fig:equivariantrd} illustrates equivariant filters initialized randomly before and after optimization.  In the two figures, each row corresponds to a different spatial scale ($J$). Since $J$ is set to 2, we have two rows. We observe that the filters in each row are the exactly the same, except for the global orientation of the wavelet.
\vspace{-10pt}

\section{No Autoaugment Ablation} 
\label{appendic:noautoaugment}
The training set of CIFAR-10 is augmented
with pre-specified autoaugment in Table \ref{table:cifarresults} to demonstrate the best possible results. To understand the effect of autoaugment, we replicate the same experiments except for not augmenting the training set with autoaugment. Table \ref{table:cifarresultsnoaa} reports the performance of the different architectures on CIFAR-10. We observe that the scattering networks followed by WRN underperform when no autoaugment is used. The difference in performance between using autoaugment and not using it is smaller when the scattering network is followed with a linear layer. 
Surprisingly, the performance of the scattering networks followed with a linear layer trained on all data increased without autoaugment. It seems that in the case of a scattering network followed by a linear model, autoaugment is not as useful as with a deep model on top and can also be harmful in some cases.  
\begin{table}[H] 
    \centering
    \caption{CIFAR-10 mean accuracy and std.\ error over 10 seeds with $J=2$ and multiple training sample sizes. The table compares the effect of augmenting the training set with pre-specified autoaugment. When the scattering network is followed by a WRN, using autoaugment is necessary to obtain better performance.  } 
    \label{table:cifarresultsnoaa}
    \small
      \fontsize{8.5}{8.5}\selectfont 
    \begin{tabularx}{340pt}{lllllll}  
        \hline
        Init. &Arch.& AA &100 samples & 500 samples & 1000 samples & All \\
        \hline
        \\[-2mm]
        TF &LS+LL$\dagger$&Yes&$37.84\pm0.57$ & $\mathbf{52.68}\pm0.31$ &$\mathbf{57.43}\pm0.17$&$69.57\pm0.1$ \\ 
        TF &LS+LL$\dagger$&No&$\mathbf{39.70}\pm0.62$ & $50.74\pm0.30$ &$54.76\pm0.22$&$\mathbf{74.94}\pm0.06$ \\ 
            \hline
        \\[-2mm]

         TF &S +LL&Yes&$36.01\pm0.55$ & $48.12\pm0.25$ &$53.25\pm 0.24$&$65.58\pm0.04$ \\ 
        TF &S +LL&No&$\mathbf{37.55}\pm0.62$ & $\mathbf{49.67}\pm0.33$ &$\mathbf{53.96}\pm 0.48$&$\mathbf{70.71}\pm0.03$ \\ 
         \hline
        \\[-2mm]
        Rand &LS+LL$\dagger$&Yes& $\mathbf{34.81}\pm0.60$ & $\mathbf{49.6}\pm0.39$ &$\mathbf{55.72}\pm0.39$&$69.39\pm0.41$ \\ 
        Rand &LS+LL$\dagger$&No& $32.64\pm0.38$ & $42.88\pm0.23$ &$47.40\pm0.32$&$\mathbf{74.71}\pm0.08$ \\  \hline
        \\[-2mm]
        Rand &S +LL&Yes& $29.77\pm0.47$ & $\mathbf{41.85}\pm0.41$ &$\mathbf{46.3}\pm0.37$&$57.72\pm0.1$ \\
        Rand &S +LL&No& $\mathbf{31.71}\pm0.34$ & $40.57\pm0.32$ &$44.42\pm0.51$&$\mathbf{61.79}\pm0.31$ \\ % &0.155530 M   \\ 
        \hline
        \\[-2mm]
         TF &LS+WRN$\dagger$&Yes&$\mathbf{43.60}\pm0.87$ & $\mathbf{63.13}\pm0.29$ &$\mathbf{70.14}\pm0.26$&$\mathbf{93.61}\pm0.12$ \\
        TF &LS+WRN$\dagger$&No&$ 34.95 \pm 0.96 $ & $ 54.21 \pm 0.39 $ &$ 62.17 \pm 0.28 $&$ 90.17 \pm 0.34$ \\ 
            \hline
        \\[-2mm]
        TF &S +WRN&Yes&$\mathbf{43.16}\pm0.78$ & $\mathbf{61.66}\pm0.32$ &$\mathbf{68.16}\pm0.27$ &$\mathbf{92.27}\pm0.05$ \\ 
        TF &S +WRN&No&$ 35.15 \pm 0.43 $ & $ 52.77 \pm 0.35 $ &$ 60.72 \pm 0.21 $ &$ 89.05 \pm 0.38$ \\
         \hline
        \\[-2mm]
        Rand &LS+WRN$\dagger$&Yes& $\mathbf{41.42}\pm0.65$ & $\mathbf{59.84}\pm0.40$ &$\mathbf{67.4}\pm0.28$&$\mathbf{93.36}\pm0.19$ \\
        Rand &LS+WRN$\dagger$&No& $ 31.08 \pm 1.00 $ & $ 48.37 \pm 0.76 $ &$ 55.41 \pm 0.49 $&$ 88.80 \pm 0.47 $ \\ 
         \hline
        \\[-2mm]
        Rand &S +WRN&Yes&$\mathbf{32.08}\pm0.46$ & $\mathbf{46.84}\pm0.21$ &$\mathbf{52.76}\pm0.33$&$\mathbf{85.35}\pm1.06$ \\
        Rand &S +WRN&No& $ 27.73 \pm 0.43 $ & $ 41.05 \pm 0.32 $ &$ 47.19 \pm 0.37 $&$ 79.67 \pm 0.59 $ \\ %
        %\hline
        %\\[-2mm]
        %&WRN&Yes& $\mathbf{38.78}\pm 0.72$ & $\mathbf{62.97}\pm 0.41$ &$\mathbf{71.37}\pm0.31$&  \\
         %&WRN& No&$ 26.33 \pm 0.49 $ & $ 47.31 \pm 0.29 $ &$ 59.85 \pm 0.41 $&$\mathbf{95.7}$^{\star}  \\ %
          % AND
        \hline
    \end{tabularx}
\begin{flushleft}
\scriptsize
\vspace{-5pt}
\hspace{85pt}$\dagger$: ours \hspace{5pt} TF: tight-frame\hspace{5pt}LS: Learnable Scattering\hspace{5pt} AA: Autoaugment\hspace{5pt}\\
\hspace{85pt}$\#$ params : 156k for S+LL; 155k for LS+LL; 11M for S+WRN; 22.6M LS+WRN; and 22.3M for WRN only\\
\end{flushleft}
% \\[-5mm]
\end{table}

\section{Cosine Loss Ablation}
\label{appendix-cosine-loss}
We replicate the experiments of Sec.~\ref{sec:results} with learnable scattering networks followed by a WRN on CIFAR-10, COVIDx-CRX2, and KTH-TIPS2. We use the same parameters except for using the cosine loss function \cite{barz2020deep} instead of cross-entropy. The cosine loss is described in Sec.~\ref{sec:relatedwork}. Wavelet filters are initialized using the tight frame construction. Table \ref{table:cifarresult_cosineloss} demonstrates the average accuracy on the three datasets. For CIFAR-10 and COVIDx-CRX2, the performance is lower when models are trained using the cosine loss. The same behavior is not observed when the models are trained on KTH-TIPS2. In fact, the performance increases slightly by using the cosine loss function. Thus, cosine loss can improve performance over small data regimes for some datasets.

\begin{table}[H] 
    \centering
    \caption{  CIFAR-10, COVIDx-CRX2 and KTH-TIPS2 mean accuracy and std.\ error using cosine loss function.} 
    \label{table:cifarresult_cosineloss}
    \small
    \fontsize{8.5}{8.5}\selectfont 
    \begin{tabularx}{400pt}{llllllll}  
        \hline
        Init. &Arch.& Dataset & Loss & 100 samples & 500 samples & 1000 samples & 1188 samples\\
        \hline
        \\[-2mm]
        TF&LS+WRN&CIFAR-10 & CE&$\mathbf{43.6}\pm0.87$ & $\mathbf{63.13}\pm0.29$&$\mathbf{70.14}\pm0.26$& -\\ 
        TF &LS+WRN& CIFAR-10 & Cosine &$42.94 \pm 0.77$ & $61.42 \pm 0.26$ & $68.29 \pm 0.18$& -\\ % &0.156096 M \\
        \hline
        \\[-2mm]
        TF &LS+WRN& COVIDx &CE &$\mathbf{81.20}\pm1.73$&$\mathbf{90.50}\pm0.70$ & $\mathbf{93.68}\pm0.35$& -\\
        TF &LS+WRN& COVIDx & Cosine &$80.03\pm2.16 $ & $89.53\pm0.89 $ & $92.75\pm0.65$& -\\
        \hline
        \\[-2mm]
        TF &LS+WRN & KTH-TIPS2 & CE &  -&-&-&$ 69.23\pm0.67 $ \\
        TF &LS+WRN & KTH-TIPS2 & Cosine &  -&-&-&$\mathbf{70.86}\pm0.67 $ \\
        \hline
    \end{tabularx}
\begin{flushleft}
\scriptsize
\vspace{-5pt}
\hspace{45pt} TF: tight-frame\hspace{5pt}LS: Learnable Scattering\hspace{5pt}S: Scattering\hspace{5pt} CE: Cross-Entropy Loss \hspace{20pt} \\
\end{flushleft}
% \\[-5mm]
\vspace{-18pt}
\end{table}

\section{Robustness to Deformations}

\label{appendix:deformations}
\begin{figure}[H]
    \centering
    \includegraphics[scale=0.21]{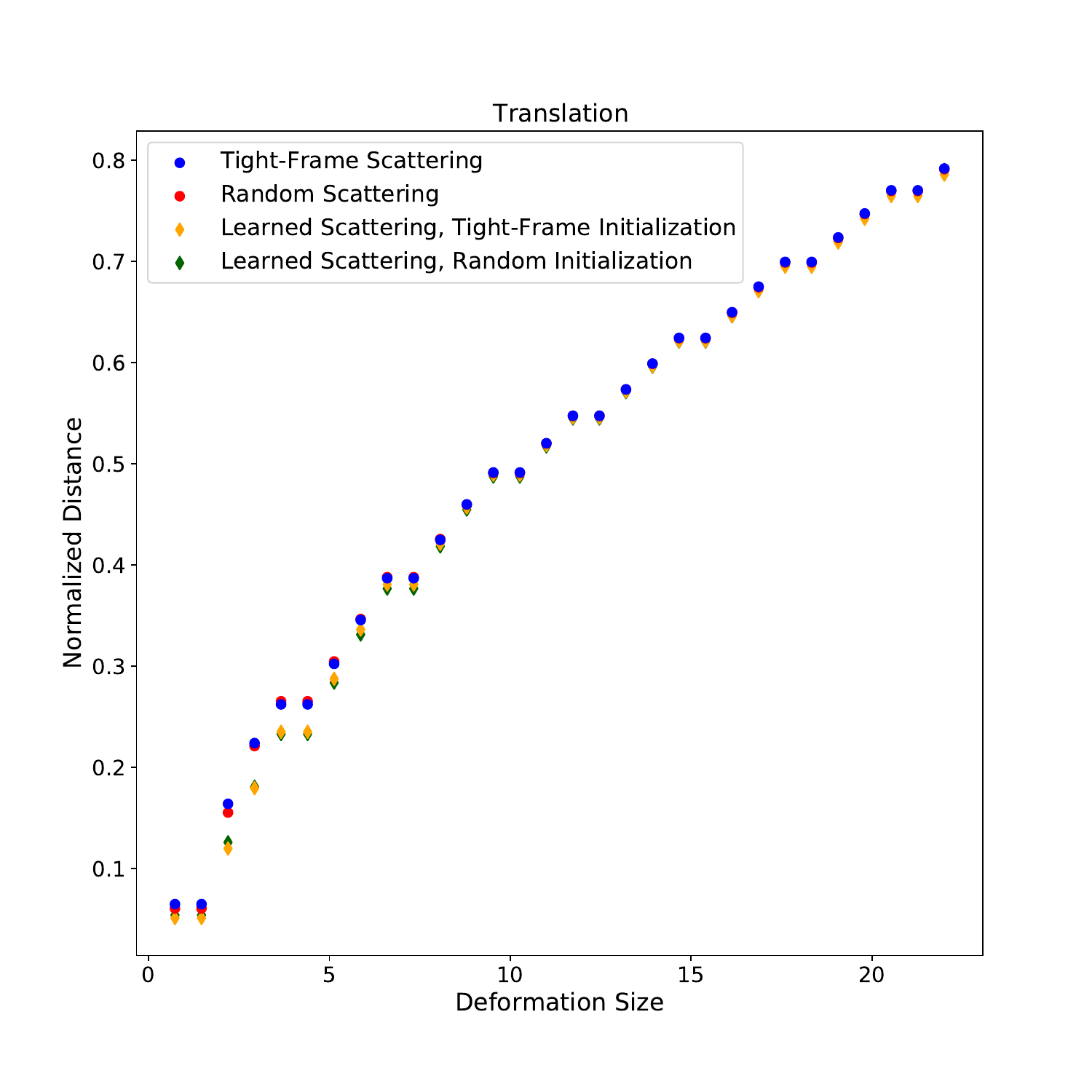}
    \includegraphics[scale=0.21]{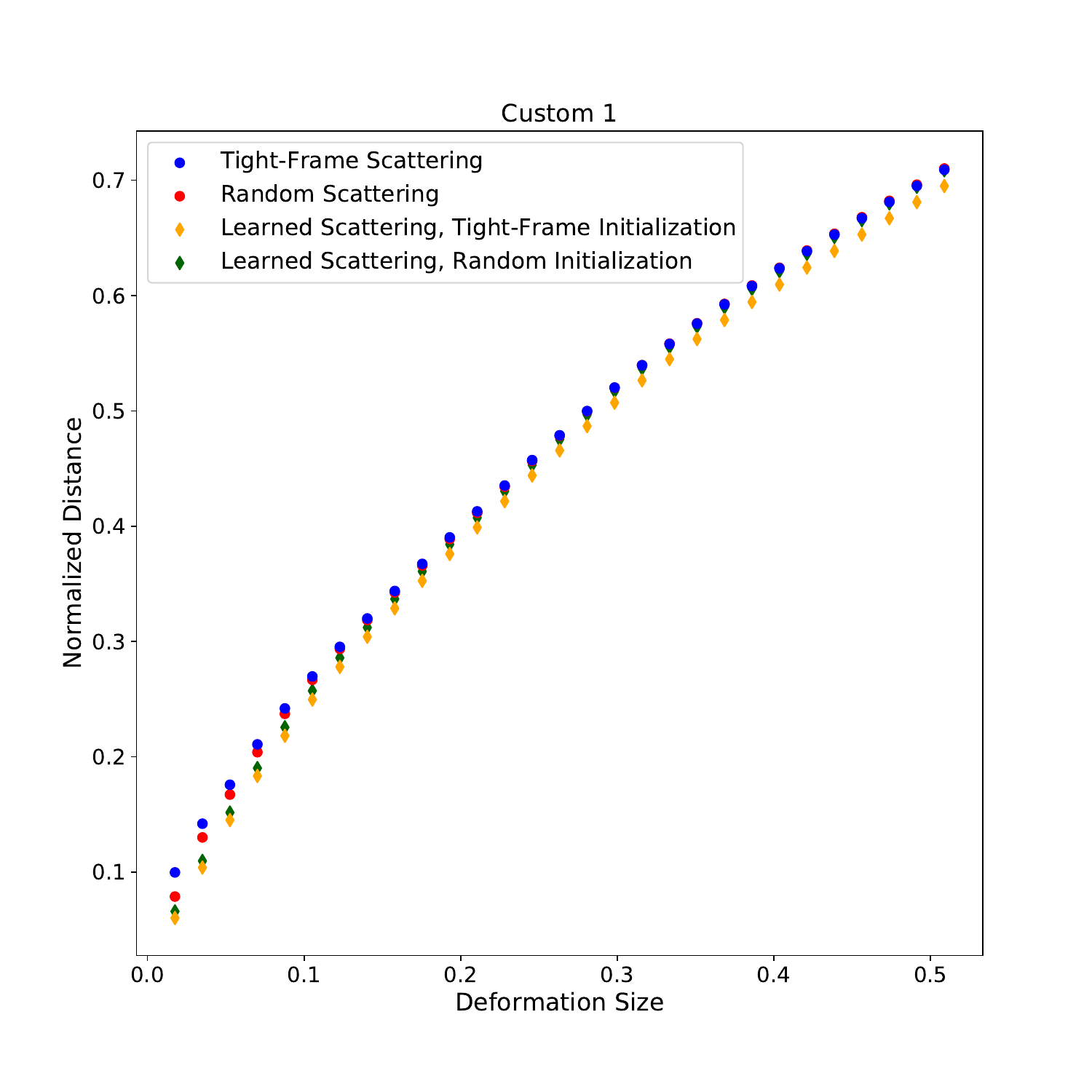}
    \includegraphics[scale=0.21]{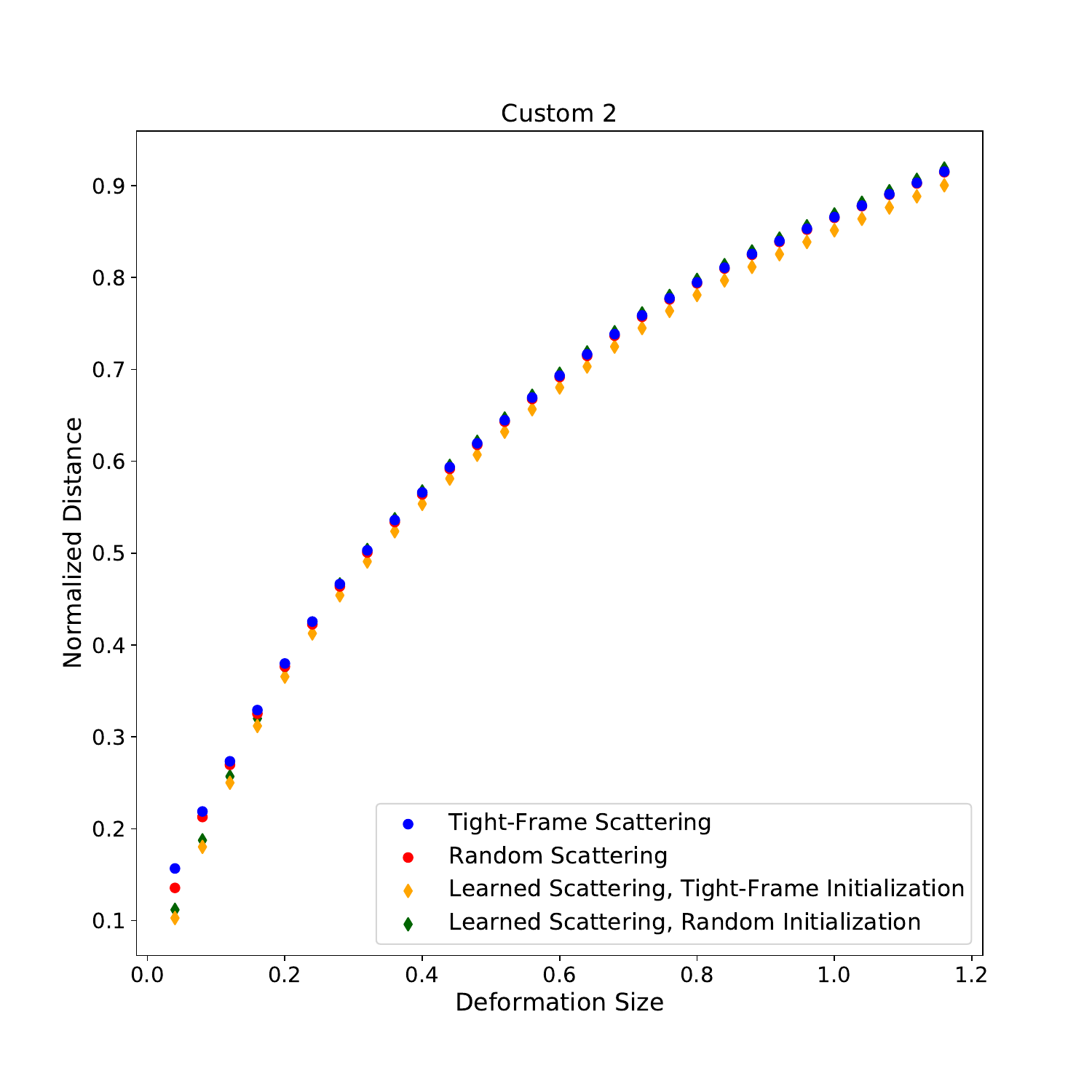}
    \vspace{-10pt}
    \caption{\textbf{Normalized distances between scattering representations of an image and its deformation.} (Left) Translation. (Middle) Custom 1 Transformation. (Right) Custom 2 Transformation. }% A comparision  Scattering representations of an image $S(x)$ and its transformation $S(\tilde{x})$   Normalized  of an X-ray image $S(x)$ and it's transformation $S(\tilde{x})$.
    \label{fig:defo-appendix}
\end{figure}

In Section~\ref{section:deformation}, we investigated if the parametric scattering transform is stable to small deformations (i.e., rotation, scale and shear). Here, we explore additional deformations. Models used were pre-trained and evaluated using the COVIDxCRX-2 setup mentioned in Section \ref{section:deformation} with a linear classifer head. The additional deformations explored are translation and several diffeomorphisms (denoted Custom 1 and Custom 2). The strengths for the all the deformations used ranges from 0 to the maximum value given in Table \ref{table:deformations}, except for scale which goes from 1 to its maximum value. Additional results are depicted in Figure \ref{fig:defo-appendix}.

Custom 1, $\tau^{1}_\epsilon(u) $, and Custom 2, $\tau^{2}_\epsilon(u)$, are defined as such:
$$
\tau^{1}_\epsilon(u) = \epsilon\begin{bmatrix}0.3u_1^2 + 0.2u_2^2 \\ 0.2(0.2u_1)\end{bmatrix}, \tau^{2}_\epsilon(u) = \epsilon\begin{bmatrix}0.3 (u_1^2 + u_2^2) \\ -0.3(2u_1 u_2)\end{bmatrix}.$$

\begin{table}[H]
    \caption{Deformations and their maximum value\vspace{-10pt}}
    \centering
    \small
    
    \label{table:deformations}
      \fontsize{10}{10}\selectfont 
    \begin{tabular}{ll}
             \hline
          Deformation & Maximum Value\\
            \hline
           Custom1 & 1 \\
           Custom2 & 1 \\
           Rotation & 10 \\
           Scale & 1.4 \\
           Shear & 5 \\
           Translation & 22 \\
\end{tabular}
\end{table}

%\textcolor{red}{TODO: LAURENT HELP ON STRENGTH PART, WTF DOES IT EXACTLY MEAN}

% SHOULD LEFT THOMAS BE MENTIONED?

\section{Details of Training Unsupervised Scattering with SimCLR Objective}
\label{appendix:unsup}
The learnable scattering networks with tight-frame and random initializations were pretrained on CIFAR-10 via the SimCLR method using a temperature of 0.5 and batch size of 128 for 500 epochs. The following basic augmentations were used as part of the SimCLR augmentation pipeline: random crop and resize, random flip, and color distortion. The optimizer used was Adam, with a learning rate of 1e-3, similar to settings from \cite{chen2020simple}. Scattering weights were then frozen and used as the backbone for a linear evaluation. For linear evaluation, we used SGD with the same optimization settings as our experiments on CIFAR-10.%, described in \ref{appendix:cifar}. 

\section{Dataset Specific parameterizations}
\label{appendix-converging}

\begin{figure}[H]
    \centering
    \includegraphics[width=\textwidth]{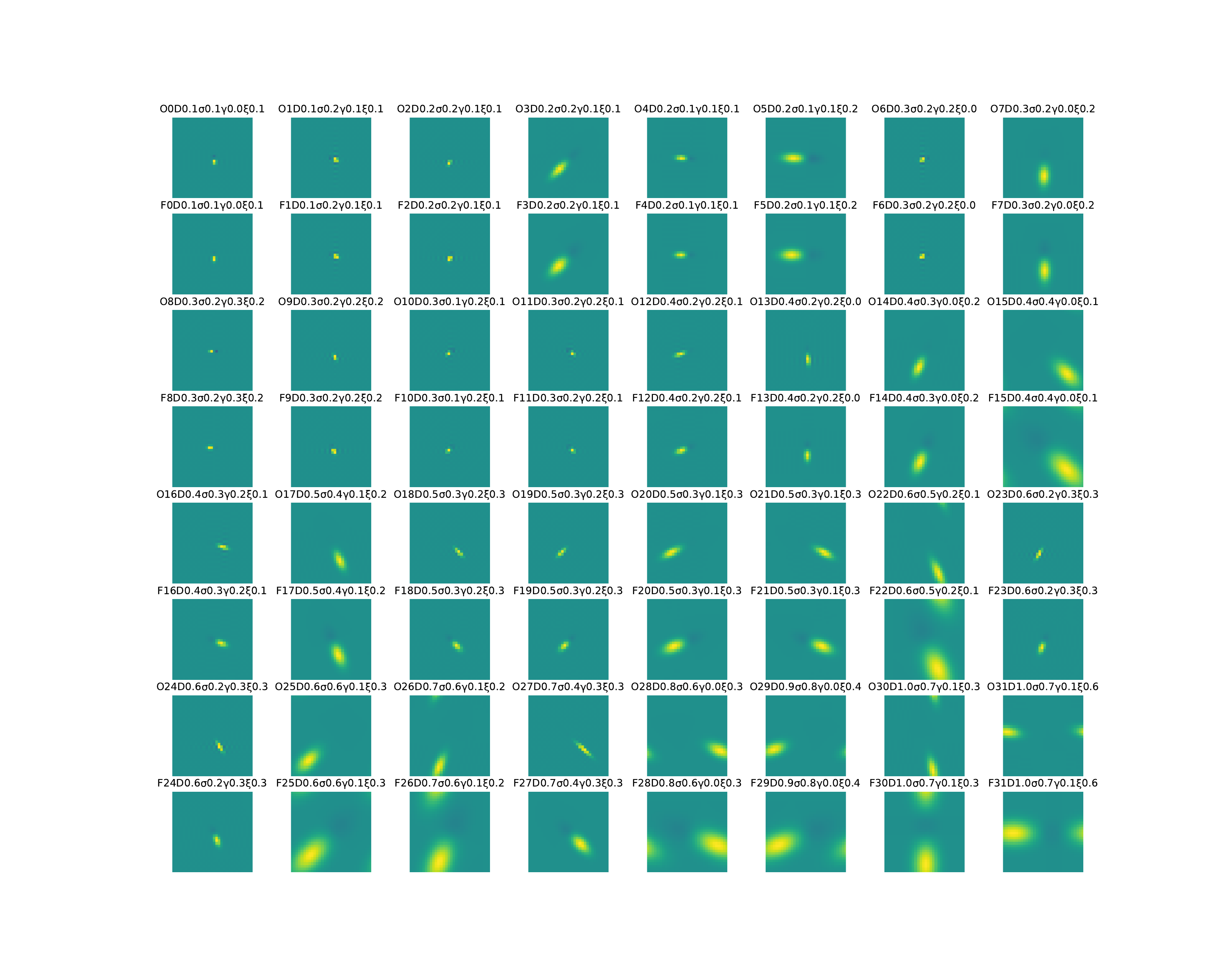}
    \vspace{-50pt}
    \caption{Filters trained on 1188 samples of KTH-TIPS2 for 500 epochs, the first, third, fifth, and seventh rows correspond to filters optimized from a tight-frame, while the second, fourth, sixth, and eighth rows correspond to tight-frame initialized filters. The filters are displayed in pairs correspond to the 'closest' (by our distance metric defined above) filters of both types. For instance, the first filter of row one matches the first filter of row two. The filters are displayed in increasing order of their distances. The top left corner corresponds to the filters that changed the least from their initialization, while the filters in the bottom right corner changed the most.}
    \label{fig:kthsorted}
\end{figure}

\begin{figure}[H]
    \centering
    \includegraphics[width=\textwidth]{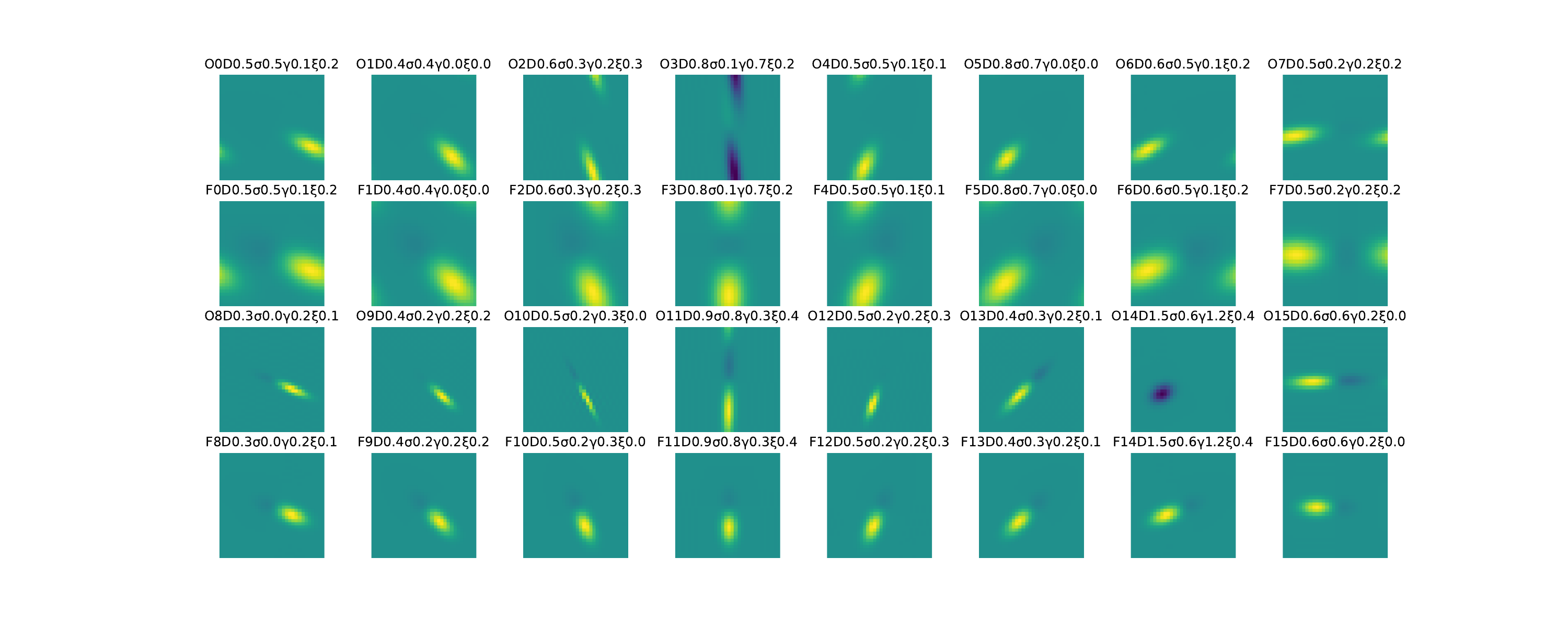}
    \vspace{-30pt}
    \caption{Filters trained on 1190 samples of CIFAR-10 for 500 epochs, the first and third,  rows correspond to filters optimized from a tight-frame, while the second and fourth rows correspond to tight-frame initialized filters. The filters are displayed in pairs correspond to the `closest' (by our distance metric defined above) filters of both types. For instance, the first filter of row one matches the first filter of row two.}
    \label{fig:cifarj}
\end{figure}

The following figures~(\ref{fig:covidj},\ref{fig:covidsorted},\ref{fig:kthj},\ref{fig:kthsorted},\ref{fig:cifarj}) show individual filter assignments obtained by hungarian matching as described in in \ref{section:converging}. They also show the individual parameter values of each filter by adopting a common naming scheme for each title. The first letter of the title is either F (fixed filters) or O (optimized filters). The next character is always a number and corresponds to the ID of the match. For all Oixxxxxxx titles, there will be a corresponding Fixxxxxxxx title; these filters correspond to the ith match. The next character is always D (distance). It is superseded by a numerical value, the Morlet wavelet distance (see \ref{section:converging}) between the filter and its match. The next character is the Gaussian window scale $\sigma$, followed by a number corresponding to the magnitude of the distances between the $\sigma$ parameters of both filters. The next character is the aspect ratio $\gamma$, followed by a number corresponding to the magnitude of the distances between the $\gamma$ parameters of both filters. The next character is the frequency scale $\xi$, followed by a number corresponding to the magnitude of the distances between the $\xi$ parameters of both filters. We omit the filter orientation $\theta$ as it is easy enough to perceive visually.
\clearpage
\subsection{COVIDX-CRX2}
\begin{figure}[H]
    \centering
    \includegraphics[width=\textwidth]{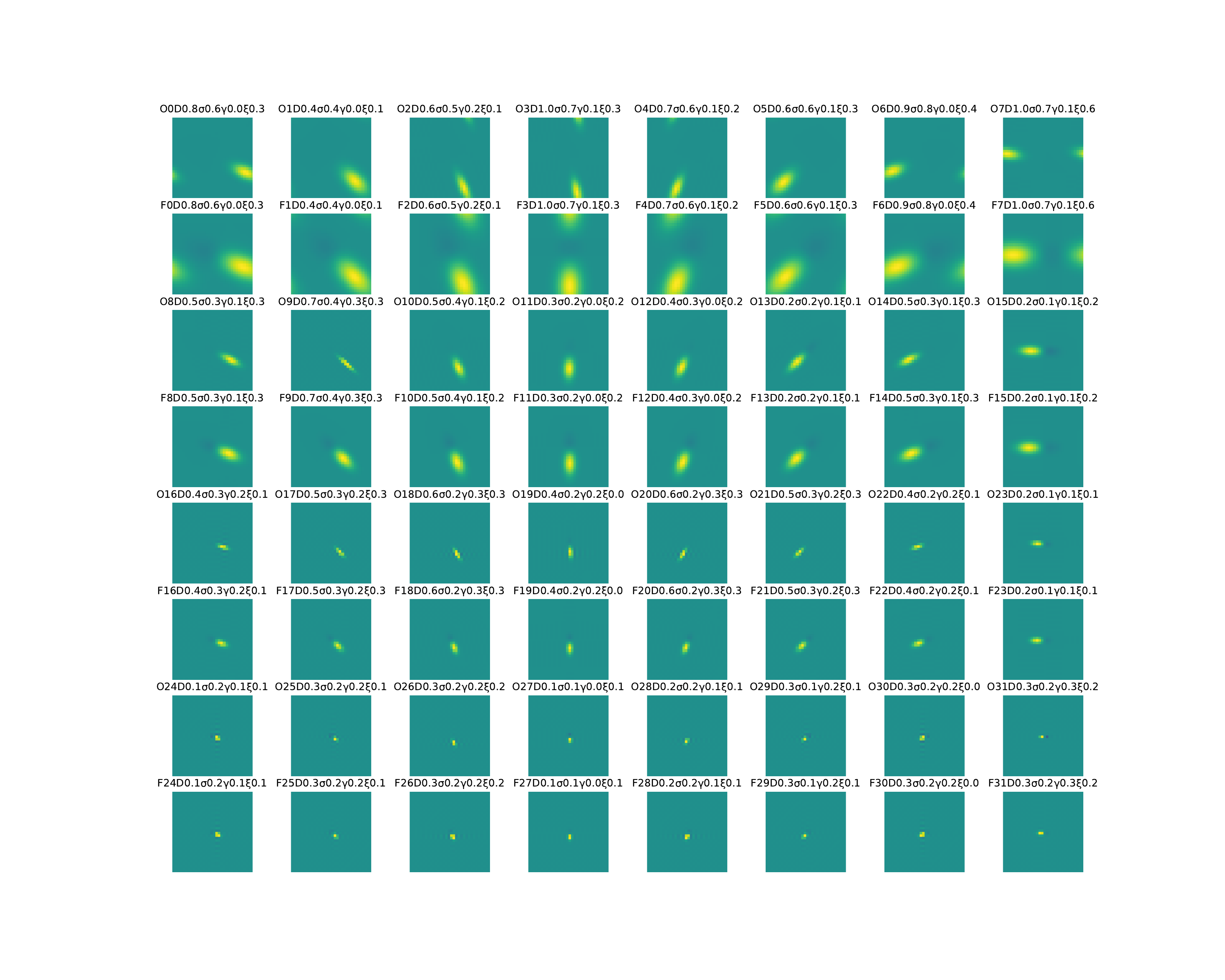}
    \vspace{-50pt}
    \caption{Filters trained on 1188 samples of COVIDx-CRX2 for 500 epochs, the first, third, fifth, and seventh rows correspond to filters optimized from a tight-frame, while the second, fourth, sixth, and eighth rows correspond to tight-frame initialized filters. The filters are displayed in pairs correspond to the `closest' (by our distance metric defined above) filters of both types. For instance, the first filter of row one matches the first filter of row 2.}
    \label{fig:covidj}
\end{figure}
\clearpage
\begin{figure}[H]
    \centering
    \includegraphics[width=\textwidth]{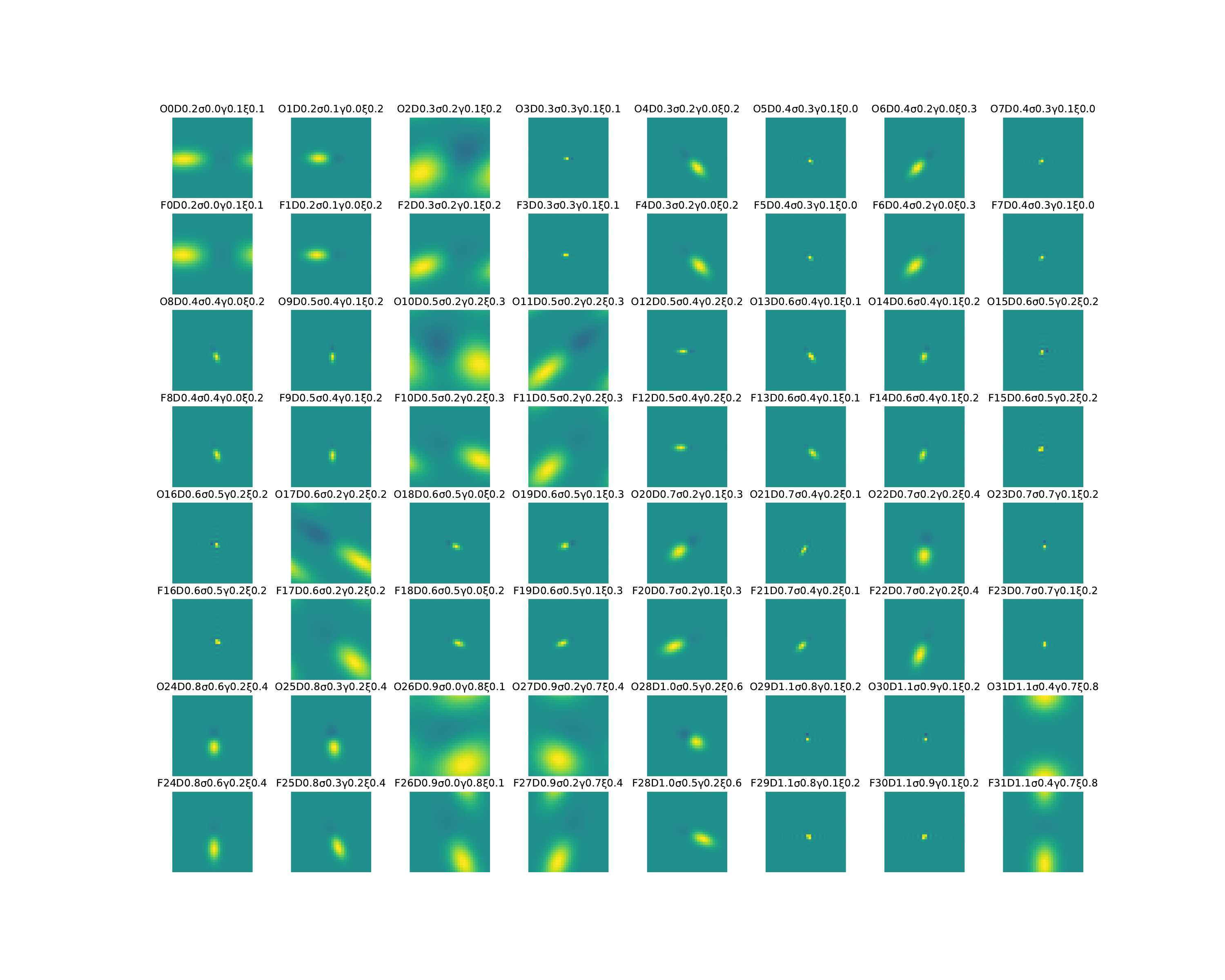}
    \vspace{-50pt}
    \caption{Filters trained on 1188 samples of COVIDx-CRX2 for 500 epochs, the first, third, fifth, and seventh rows correspond to filters optimized from a tight-frame, while the second, fourth, sixth, and eighth rows correspond to tight-frame initialized filters. The filters are displayed in pairs correspond to the `closest' (by our distance metric defined above) filters of both types. For instance, the first filter of row one matches the first filter of row 2. The filters are displayed in increasing order of their distances. The top left corner corresponds to the filters that changed the least from their initialization, while the filters in the bottom right corner changed the most.}
    \label{fig:covidsorted}
\end{figure}

\clearpage
\subsection{KTH-TIPS2}
\begin{figure}[H]
    \centering
    \includegraphics[width=\textwidth]{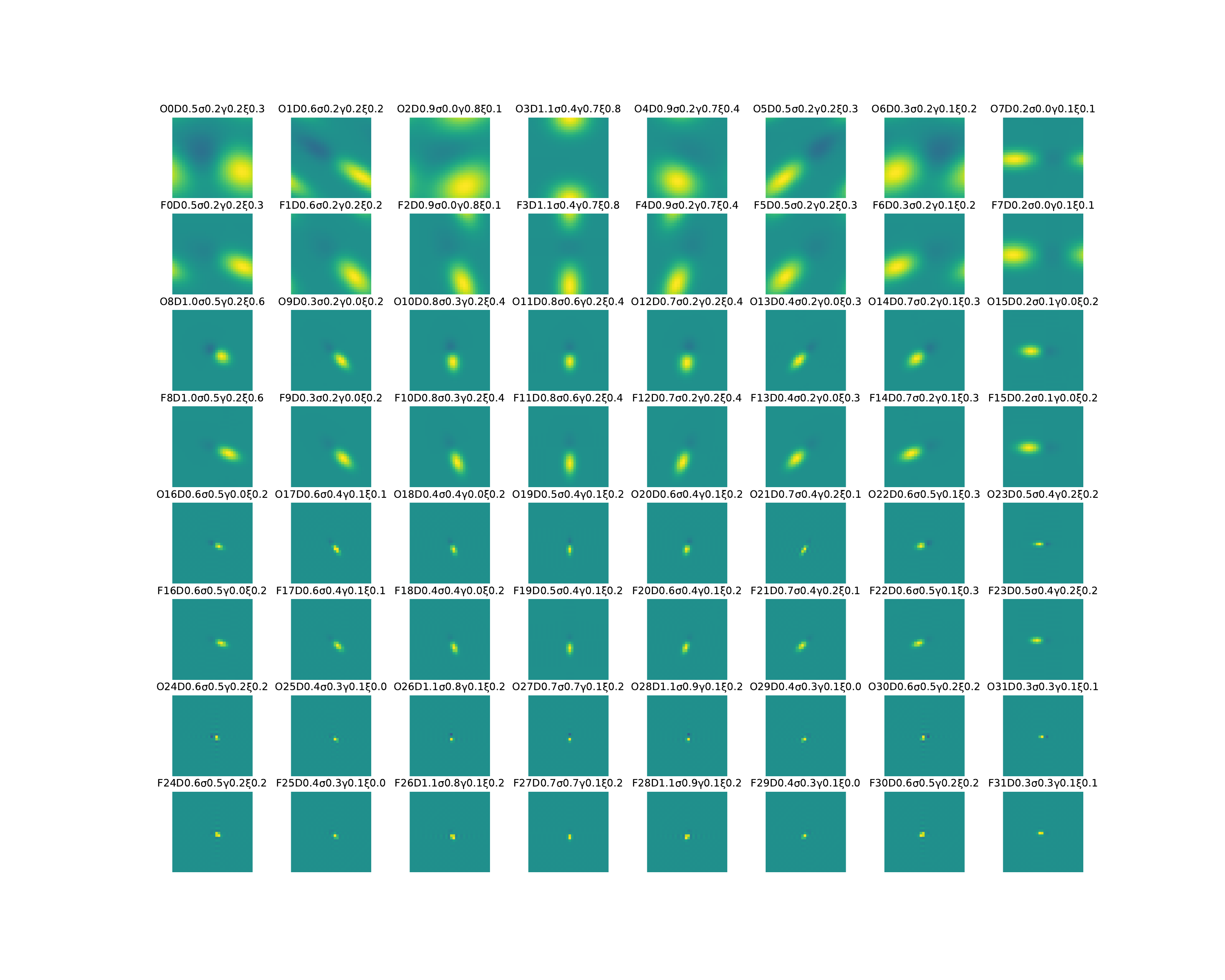}
    \vspace{-50pt}
    \caption{Filters trained on 1188 samples of KTH-TIPS2 for 500 epochs, the first, third, fifth, and seventh rows correspond to filters optimized from a tight-frame, while the second, fourth, sixth, and eighth rows correspond to tight-frame initialized filters. The filters are displayed in pairs correspond to the `closest' (by our distance metric defined above) filters of both types. For instance, the first filter of row one matches the first filter of row two.}
    \label{fig:kthj}
\end{figure}
\clearpage

\subsection{CIFAR-10}

\subsection{Dataset Specific initializations with a Random Initialization}\label{appendix:random-init}
\begin{figure}[H]
    \centering
\includegraphics[width=0.7\textwidth]{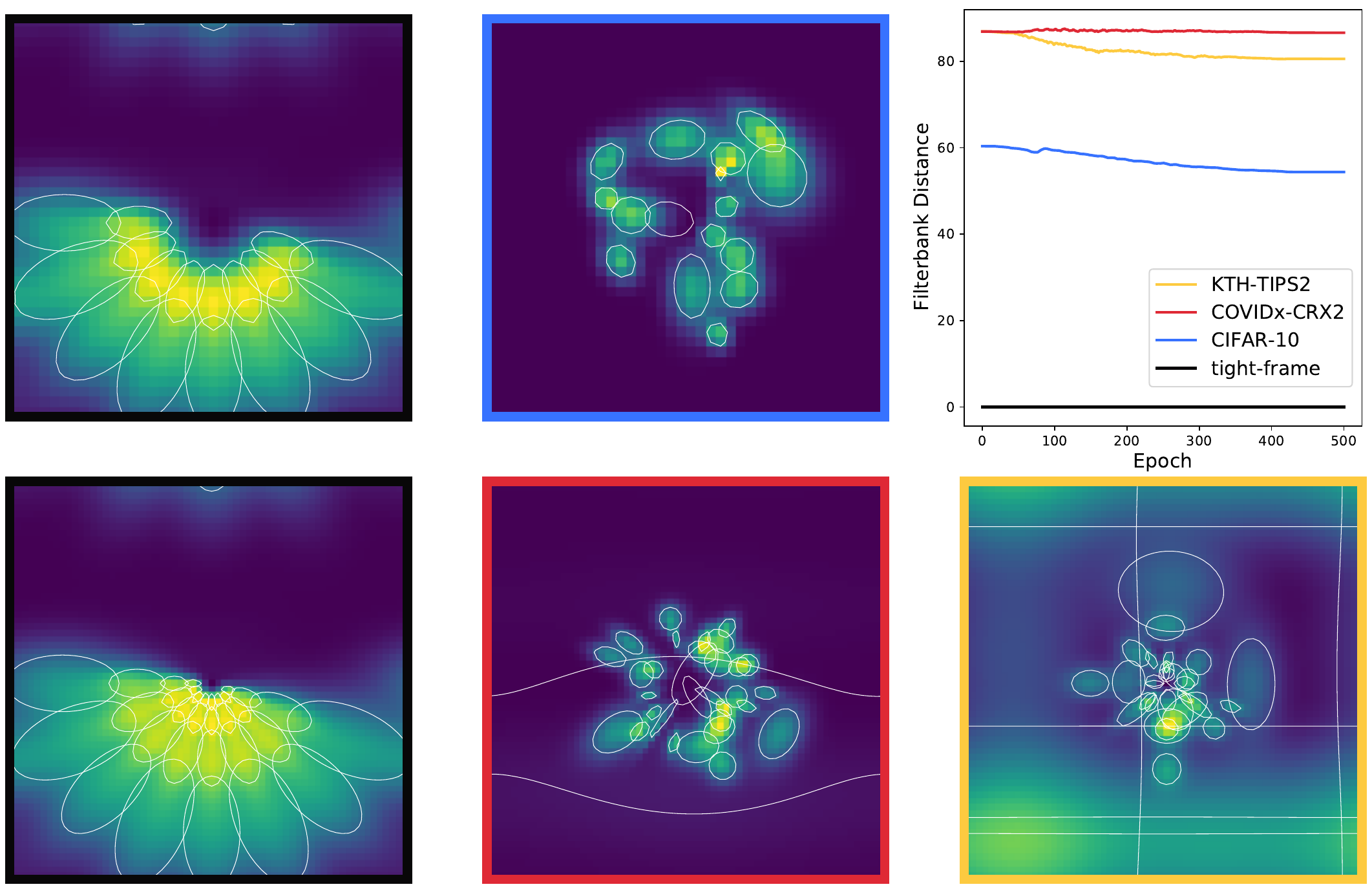}
    \vspace{-7pt}
    \caption{The graph shows the \textit{filterbank distance} over epochs as the filters, initialized from random initialization, are trained on different datasets. To the left, we visualize dataset specific parameterizations of scattering filterbanks in Fourier space. %White contours are drawn around each morlet wavelet for clarity. 
    The graph on the right shows that the randomly initialized filterbanks become more similar to a tight frame during training.}
    \label{fig:graphlwp_random}
\end{figure}

In Figure~\ref{fig:graphlwp_random}, we show how the filters adapt when initialization begins from a random setting. We note the deviation to tight frame is much greater than in the case where we initialize in a tight frame. However, as per our filterbank distance, we observe the filters do move closer to the tight frame than their initialization.

\section{Number of Filters Ablation}
\label{appendix:num_filter}
In this section, we investigate the effect of modifying the number of filters ($L$) per spatial scale on CIFAR-10. We use the same settings and hyperparameters as described in Appendix~\ref{appendix-details}. So far, in all experiments, we have set the number of filters per scale at 8 for fair comparison with parameters.

For this ablation, we train a parametric scattering network followed by a linear layer where the wavelets are initialized using the tight frame construction and the canonical parameterization. The spatial scale is set to 2. We do not use autoaugment on the training set since, as shown in Appendix~\ref{appendic:noautoaugment}, autoaugment can be harmful when the scattering network is followed by a linear layer. Table~\ref{table:L} shows the accuracy of the full training set for different values of $L$. We observe that the performance increases when the number of filters per scale also increases. Around 14-16 filters per spatial scale, the performance seems to have stopped increasing. 

Table~\ref{table:Lcompare} demonstrates the mean accuracy over different sizes of training samples where the number of filters per scale is set to 8 or 16. We also consider the fixed version of the scattering transform. Over all training sample sizes, we observe that the highest performance is obtained with learnable scattering using 16 filters per spatial scale. When the scattering network is fixed, we observe that performances are lower with 16 filters instead of 8. It seems that increasing the number of filters per scale is only beneficial in the learned version of the scattering network.

\begin{table}[H] 
    \centering
    \caption{CIFAR-10 accuracy of learnable scattering followed by a linear layer (LS + LL) and multiple numbers of filters per scale ($L$) trained on all the training set. The wavelet filters are initialized using the tight frame construction and the canonical parameterization. The spatial scale is set to 2. No autoaugment is used for this experiment. We observe that the performance increases when the number of filters per scale ($L$) also increases. Around 14 filters per spatial scale, the performance seems to have stopped increasing. } 
    \label{table:L}
    \fontsize{9}{9}\selectfont 
    \begin{tabularx}{0.15\linewidth}{ll} 
          \hline
         L & All Data\\
        \hline
        \\[-2mm]
        2& $63.59$\\
        4& $70.94$\\
        6& $74.03$\\
        8& $74.94$\\
        10& $76.40$\\
        12& $77.01$\\
        14& $\mathbf{77.36}$\\
        16& $77.33$\\
        \hline
    \end{tabularx}
\end{table} 
\begin{table}[H] 
    \centering
    \caption{CIFAR-10 mean accuracy and std.\ error over 10 seeds with multiple training sample sizes and different values of $L$. The wavelet filters are initialized using the tight frame construction. The spatial scale is set to 2. No autoaugment is used for this experiment. Over all training sample sizes, we observe that the highest performance is obtained with learnable scattering using 16 filters per spatial scale. }
    \label{table:Lcompare}
    \fontsize{9}{9}\selectfont 
    \begin{tabularx}{0.75\linewidth}{lllllll} 
          \hline
        Arch. &Parameterization & L & 100 samples &500 samples & 1000 samples & All \\
        \hline
        \\[-2mm]
        LS+LL$\dagger$&Canonical& 8&$39.70\pm0.62$ & $50.74\pm0.30$ 
        &$54.76\pm0.22$&$74.94$\\ 
        LS+LL$\dagger$&Canonical& 16& $39.73\pm0.39$& 
        $\mathbf{54.17}\pm0.36$ &$\mathbf{58.36}\pm0.29$&$\mathbf{77.33}$ \\ 
        S +LL&-&8 &$37.55\pm0.62$ & $49.67\pm0.33$ &$53.96\pm 0.48$&$70.71$ \\
        S +LL&-&16 & $35.85 \pm 0.48$	 & $48.2 \pm 0.27$  & $52.74 \pm 0.25$ &$70.64$ \\
        \hline
    \end{tabularx}
\end{table}

\section{Perturbing a Converged Parameterization Ablation}

\begin{figure*}[ht]
    \centering
    \includegraphics[width=0.45\textwidth]{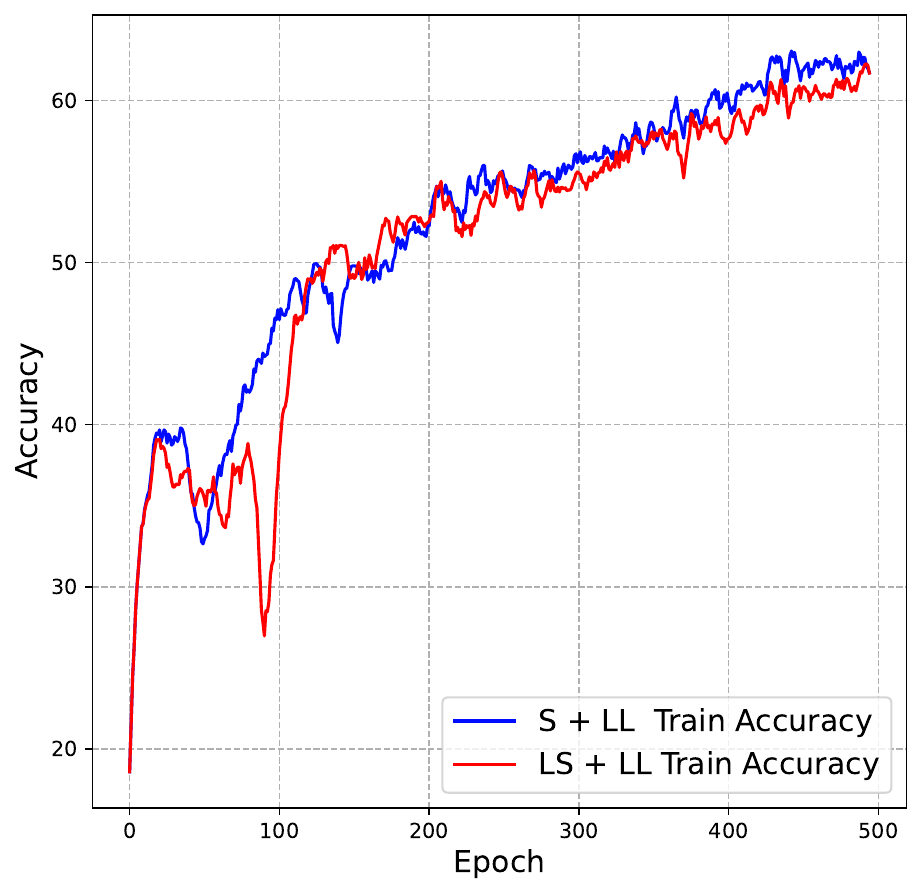}%
    \qquad
\includegraphics[width=0.45\textwidth]{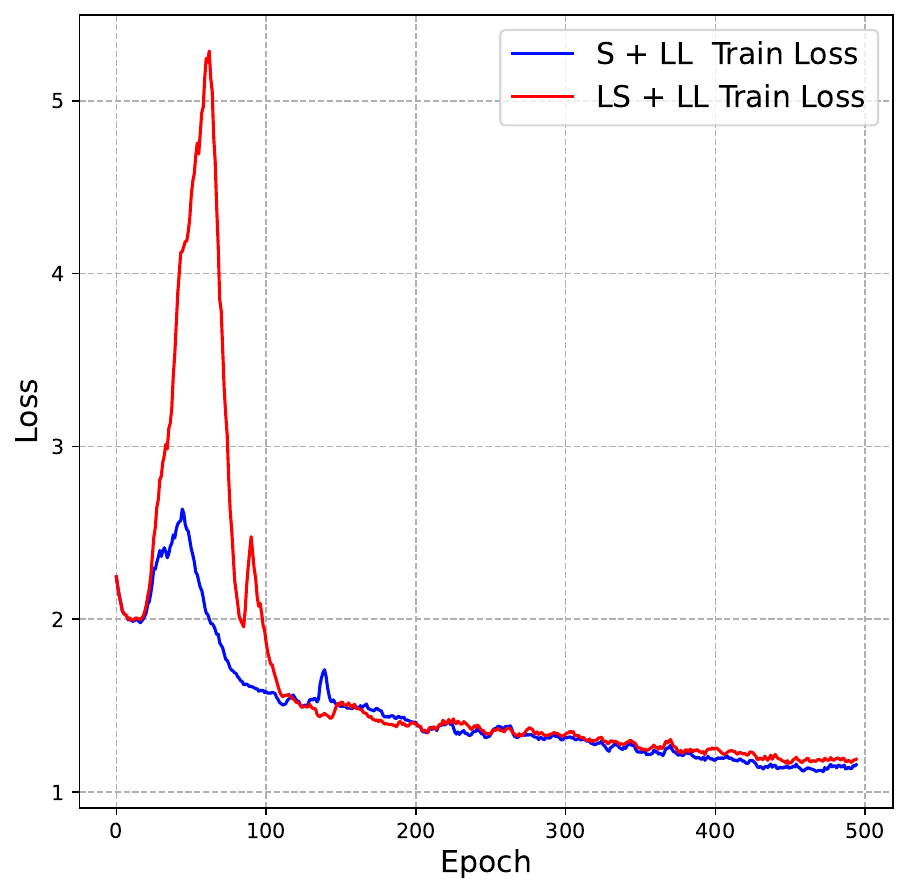} %
\caption{Plots comparing the training accuracies and losses over epochs of LS+LL and S+LL. Note that scattering parameters are only optimized for LS+LL. The networks were trained for 500 epochs on 1000 samples of CIFAR-10.} %We note that training statistics were taken each epoch and smoothed over a window of 25, while test statistics were taken at intervals of 25 epochs and smoothed over a window of 5.}
\label{fig:learningcurve}
\end{figure*}

\begin{figure}
\noindent\adjustbox{width=\columnwidth}{
         \begin{tabular}{ccc|ccc}
         \textbf{Initial} & \textbf{Perturbed} & \textbf{Final} & \textbf{Initial} & \textbf{Perturbed} & \textbf{Final}\\
         \adjustbox{width=1.8cm}{\includegraphics[trim=2.3in 1.3in 2.1in 1.4in, clip]{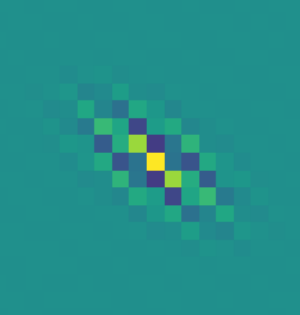}}& 
         \adjustbox{width=1.8cm}{\includegraphics[trim=2.3in 1.3in 2.1in 1.4in, clip]{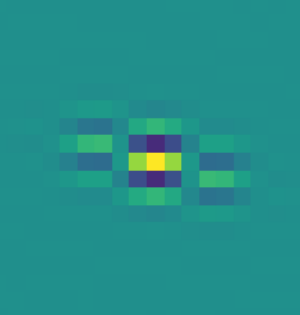}} & 
         \adjustbox{width=1.8cm}{\includegraphics[trim=2.3in 1.3in 2.1in 1.4in, clip]{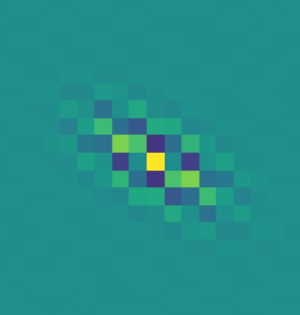}} &
         \adjustbox{width=1.8cm}{\includegraphics[trim=2.3in 1.3in 2.1in 1.4in, clip]{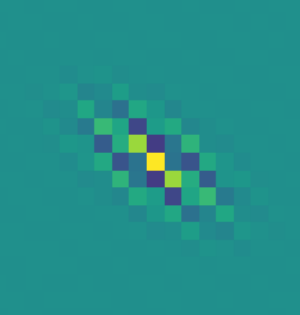}}& 
         \adjustbox{width=1.8cm}{\includegraphics[trim=2.3in 1.3in 2.1in 1.4in, clip]{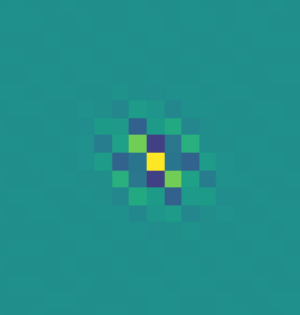}} & \adjustbox{width=1.8cm}{\includegraphics[trim=2.3in 1.3in 2.1in 1.4in, clip]{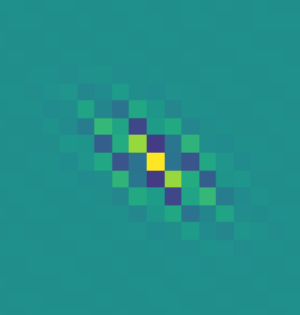}} \\  
\end{tabular}}
    \caption{(left) $\theta$  parameter of a learned parameterization is perturbed and returns to an almost identical orientation after gradient based optimization. (right) $\xi$ parameter of a learned parameterization is perturbed and returns to an almost identical frequency scale after gradient based optimization. }
    \label{fig:pert1}
\end{figure}

\begin{figure}
\centering
\small
\noindent\adjustbox{width=\columnwidth}{
         \begin{tabular}{ccc|ccc}
         \textbf{Initial} & \textbf{Perturbed} & \textbf{Final} & \textbf{Initial} & \textbf{Perturbed} & \textbf{Final}\\
         \adjustbox{width=1.8cm}{\includegraphics[trim=2.3in 1.3in 2.1in 1.4in, clip]{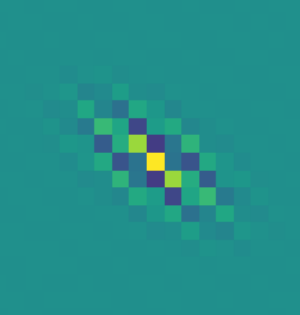}}& 
         \adjustbox{width=1.8cm}{\includegraphics[trim=2.3in 1.3in 2.1in 1.4in, clip]{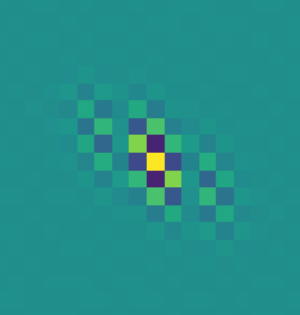}} & 
         \adjustbox{width=1.8cm}{\includegraphics[trim=2.3in 1.3in 2.1in 1.4in, clip]{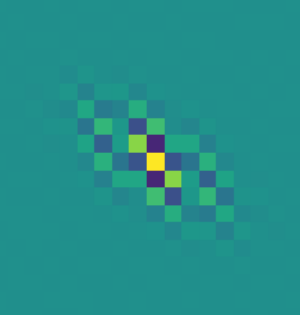}} &
         \adjustbox{width=1.8cm}{\includegraphics[trim=2.3in 1.3in 2.1in 1.4in, clip]{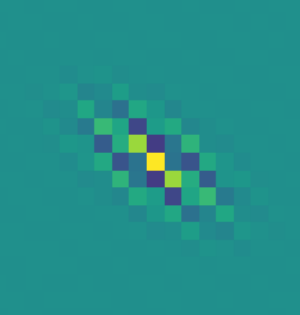}}& 
         \adjustbox{width=1.8cm}{\includegraphics[trim=2.3in 1.3in 2.1in 1.4in, clip]{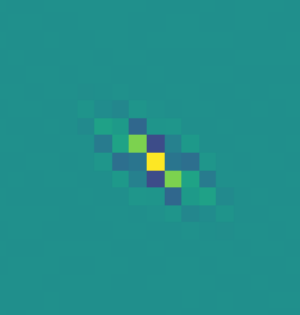}} & \adjustbox{width=1.8cm}{\includegraphics[trim=2.3in 1.3in 2.1in 1.4in, clip]{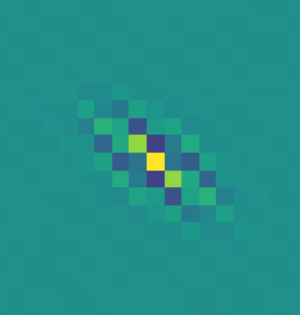}} \\ 
\end{tabular}}
    \caption{(left) $\sigma$  parameter of a learned parameterization is perturbed and returns to an almost identical gaussian window scale after gradient based optimization. (right) $\gamma$ parameter of a learned parameterization is perturbed and returns to an almost identical aspect ratio after gradient based optimization.}
    \label{fig:pert2}
\end{figure}

The training accuracies and losses over epochs are shown in Figure~\ref{fig:learningcurve} for LS+L and S+L. We observe that both networks are following similar patterns. Local minima are expected here (see results with different init Figure~\ref{fig:filters-cifar-real-before-kymatio}), as in any non-convex problem. To showcase learnable scattering's stability to stochastic optimization, we perform an ablation below (Figures~\ref{fig:pert1} \& \ref{fig:pert2}) for trained LS+L, where each parameter is perturbed and then individually optimized. We find that they all return rapidly to the (local) optimum.

% In practise, we found standard stochastic optimizers popular in DL to be stable in our case, with training curves (Fig. \ref{fig:learningcurve}) following similar patterns for LS+L and S+L. Local minima are expected here (see results with different init Fig. 1), as in any non-convex problem. To showcase learnable scattering's stability to stochastic optimization, we perform an ablation below (Figs \ref{fig:pert1} \& \ref{fig:pert2}) for trained LS+L, where each parameter is perturbed and then individually optimized, finding they all return rapidly to the (local) optimum (e.g., the two cases below). 

% \begin{figure*}[H]
%     \centering
%     \subfloat[\centering Accuracy: WRN-16- v.s. LS+WRN ]{{\includegraphics[width=0.45\textwidth]{img/learning_curves/accuracy_onlycnn_lscnn.pdf} }}%
%     \qquad
% \subfloat[\centering Loss: WRN-16- v.s. LS+WRN]{{\includegraphics[width=0.45\textwidth]{img/learning_curves/loss_onlycnn_lscnn.pdf} }}%
% \caption{Plots comparing WRN-16- and LS+WRN learning curves are shown for 3000 epochs of training on 100 saples of CIFAR-10. We note that training statistics were taken each epoch and smoothed over a window of 25, while test statistics were taken at intervals of 25 epochs and smoothed over a window of 5. }
% \label{fig:qlearningOverEpisodes}
% \end{figure*}

\clearpage
\section{Visualizing Parameter Values over Time}

\begin{figure}[H]
    \centering
    \includegraphics[width=\textwidth]{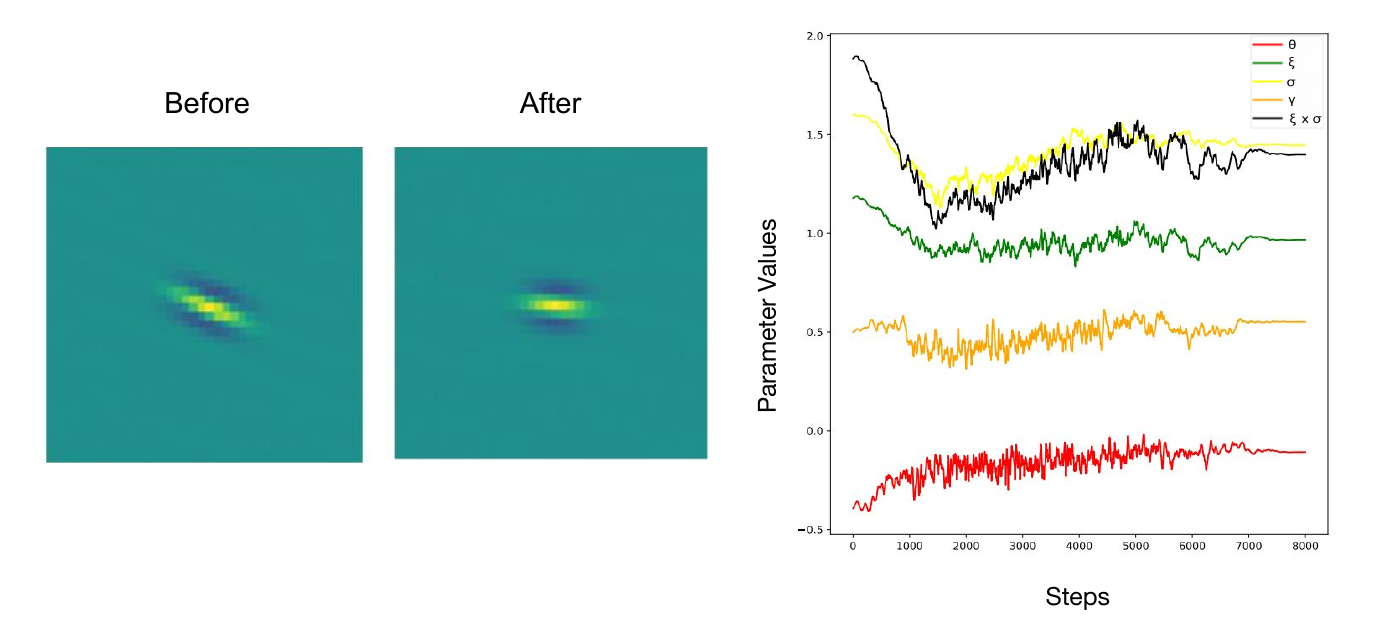}
    \caption{Parameter values over time. Real part of Morlet wavelet filters initialized with \textit{tight-frame}  schemes before (left) and after (middle) training. The network was trained for 1K epochs on 1000 samples of CIFAR-10. We use the Morlet canonical wavelet parameterization. The plot (right) shows the parameter values over epochs. }
    \label{fig:params_over_time}
\end{figure}

 Figure~\ref{fig:params_over_time} shows, on the left, a Morlet wavelet filter before training and, in the middle, the same wavelet filter after training. The parametric scattering network was trained for 1K epochs on 1000 samples of CIFAR-10. We use the Morlet canonical parameterization. We observe that the global orientation has changed during training. Figure~\ref{fig:params_over_time} (right) shows the values of the parameters from Table~\ref{tab:params} over the training steps. We observe that the value of all parameters changed during training. However, the aspect ratio seems to have returned to its initial value. In Figure~\ref{fig:params_over_time} (right), we also show the value of $\xi \times \sigma$, which is a proxy for the wavelet scale factor, over the steps. The value of $\xi \times \sigma$ is smaller after training. This can be observed in Figure~\ref{fig:params_over_time} (left-middle), where the wavelet after training seems to be more compressed than before training.

\end{document}